\documentclass{WileyMSP-template}

\usepackage{siunitx}
\DeclareSIUnit\mmHg{mmHg}
\DeclareSIUnit\bpm{bpm}
\usepackage{soul}
\usepackage{subcaption}
\usepackage{url}
\usepackage{microtype}
\usepackage[spanish, english]{babel}

\newcommand{\cites}[1]{\textsuperscript{\cite{#1}}}

\begin{document}
\rhead{\includegraphics[width=2.5cm]{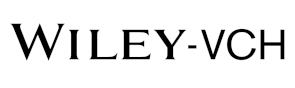}}

\title{An Integrated Soft Robotic System for Measuring Vital Signs in Search and Rescue Environments}

\maketitle


\author{Jorge Francisco García-Samartín*}
\author{Dr. Christyan Cruz Ulloa}
\author{Andrés Sánchez-Silva}
\author{Dr. Jaime del Cerro}
\author{Prof. Dr. Antonio Barrientos*}



\begin{affiliations}
García-Samartín, J.F., Barrientos, A.\\
Centro de Automática y Robótica (UPM - CSIC), Universidad Politécnica de Madrid, Calle José Gutiérrez Abascal 2, 28006, Madrid, Spain\\
Email Address: jorge.gsamartin@upm.es, antonio.barrientos@upm.es

Cruz Ulloa, C., Sánchez-Silva, A., del Cerro, J.\\
Centro de Automática y Robótica (UPM - CSIC), Universidad Politécnica de Madrid, Calle José Gutiérrez Abascal 2, 28006, Madrid, Spain

\end{affiliations}



\begin{abstract}

Robots are frequently utilized in search-and-rescue operations. In recent years, significant advancements have been made in the field of victim assessment. However, there are still open issues regarding heart rate measurement, and no studies have been found that assess pressure in post-disaster scenarios. This work designs a soft gripper and integrates it into a mobile robotic system, thereby creating a device capable of measuring the pulse and blood pressure of victims in post-disaster environments. The gripper is designed to envelop the victim's arm and inflate like a sphygmomanometer, facilitated by a specialized portability system. The utilization of different signal processing algorithms has enabled the attainment of a pulse bias of \qty{4}{\bpm} and a bias of approximately \qty{5}{\mmHg} for systolic and diastolic pressures. The findings, in conjunction with the other statistical data and the validation of homoscedasticity in the error terms, prove the system's capacity to accurately determine heart rate and blood pressure, thereby rendering it suitable for search and rescue operations. Finally, a post-disaster has been employed as a test to validate the functionality of the entire system and to demonstrate its capacity to adapt to various victim positions, its measurement speed, and its safety for victims.

\end{abstract}


\section{Introduction}

Robotics is emerging as a critical component in search and rescue (SAR) operations. In the context of escalating natural disasters, such as earthquakes, floods, and wildfires, as well as man-made catastrophes, the use of robots has become a strategy to mitigate the exposure of human rescue teams to extreme risks, such as structural instability, toxic leaks, and unpredictable environmental hazards\cites{Delmerico2019}. Consequently, a variety of solutions have been developed, both for individual robots\cites{Meenakshi2025} and collaborative swarms\cites{Wang2023Development}, for tasks as varied as the exploration of post-disaster scenarios\cites{Sampedro2018A, Queralta2020}, hazard assessment\cites{CruzUlloa2024}, victim search and recognition\cites{CruzUlloa2023, Garcia-Samartin2024b}, and even physical intervention\cites{Schwaiger2024UGV}.

While significant advancements have been observed in these domains, with ample opportunities for further development, similar progress has not been witnessed in the field of victim assessment. Nonetheless, this task remains of significant importance. Ascertaining the condition of the victim, especially during the first hour\cites{newgard2010emergency}, is of paramount importance in order to facilitate the triage~\cites{Zhang2023c}. Regardless of whether this process is executed automatically\cites{Chong2016} or, as is still commonplace in the present day, by human operators\cites{Petricia2026}, it is imperative to prioritize and establish rescue routes with the objective of maximizing the number of lives that can be saved and, when it is the case, minimizing the time that rescuers spend in the post-disaster scenario.

In the medical field, four parameters are considered essential when assessing the urgency of care for a victim: temperature, respiratory rate, pulse, and blood pressure. Of these, the determination of the first has been achieved with errors lower than \qty{1}{\celsius}\cites{Gong2020SHUYURobot, Huang2022MobileRobotic}, while the second is currently the focus of the most research. This is because breathing produces movement.

The most commonly measured magnitude using robotic systems has been breathing frequency because it produces significant movement that can be captured by different sensors. Although radars can monitor this vital sign even when victims are obstructed by obstacles such as doors, walls, or debris, they have an error rate of 10\%\cites{Schroth2024}. This error rate can only be slightly reduced through extensive signal processing\cites{Jing2025Advancing}. In contrast, other alternatives, such as RGB cameras\cites{Huang2019, Huang2020, Orbea2024} or laser sensors\cites{Cittadini2024Contactless}, produce errors of 2\% or \qty{2}{\bpm}, while Dou et al.'s work\cites{Dou2021}, where the respiration rate is gauged by examining the impact of the Doppler effect on Wi-Fi signals, reaches errors of less than 0.15 bpm.

Nevertheless, different studies suggest that determining heart rate and arterial pressure yields more robust results\cites{Wangdi2025-zs} and allows one to determine the hemodynamic state of victims, consequently enabling to identify cases of shock, hypovolemia, or hypoperfusion\cites{Eichlseder2025}. Moreover, unlike breathing frequency, which tends to remain stable, both magnitudes evolve over time and enable identification of severe damage\cites{lavery2007utility, mutschler2013shock}. Because robots can measure vital signs without risk over long periods of time, measuring heart rate and arterial pressure is very useful in such situations.

Robots that can measure the heart rates of victims have been developed. They typically use visual photoplethysmography (PPG) to study skin color variations and achieve mean square errors over \qty{10}{\bpm}\cites{Huang2022MobileRobotic, Jing2025Advancing}, though there are robotic platforms ranging from 20\cites{Pająk2023Touchless} to \qty{2}{\bpm}\cites{Huang2020}, errors considered sufficient given that the aim is to assess the victim's condition, not to perform a clinical examination . Although there is little literature on using radar to measure heart rate, the accuracy appears to be similar\cites{Hussein2023Contactless}. The tactile sensor of\cites{Kerr2021}, on its side, gives errors lower than \qty{2}{\bpm}.

Conversely, no robots have been identified that measure pressure in the SAR field. The utilization of automated medical devices for blood pressure measurement has been observed in controlled environments, suggesting a promising future for this research domain.

Specifically, two lines exist in this field. On the one hand, researchers utilize contact sensors in proximity to subcutaneous blood vessels to obtain a PPG. Following a series of signal treatments, they estimate blood pressure. An origami finger was developed in\cites{Kim20223D}. The pressure is derived from analytical relations. Although the device appears to be effective, it has only been tested on a single subject. Furthermore, the device must be applied to the chest, which can be easily occluded by clothing.

In a similar vein, the work of Cen et al.\cites{Cen2023OCViTnet} employs a transformer-based deep learning model to extract blood pressure from cardiac signals. The objective of the study is to integrate the sensor on a robot with tactile sensors that are prepared to be applied on a human wrist. However, the testing process has been conducted using preexisting datasets, not those collected by the system. The results of these tests have shown that there are MAE errors of \qty{7.85}{\mmHg} for systolic pressure and \qty{5.52}{\mmHg} for diastolic pressure. Despite the fact that it does not directly measure blood pressure, the pulse pressure index tonometry pulse sensor of\cites{Jun2020Accuracy} has been shown to achieve accuracies of 98\% by employing an array of piezoresistive sensors and classical signal treatment.

An alternative method for measuring pressure involves the use of visual PPG. This approach has been previously explored in\cites{Maher2021Design} and\cites{Chiang2022Contactless}. In the first of both works, a variety of machine learning algorithms were integrated into a mobile robot. A MAE of \qty{4.77}{\mmHg} has been reported for systolic pressure and \qty{4.8}{\mmHg} for diastolic pressure. However, the methodology related to the tests is not described. In the second study, a machine learning model capable of estimating blood pressure is developed. To this end, they employed a CNN, which yielded a MAE of \qty{2.27}{\mmHg} for systolic pressure and \qty{0.8}{\mmHg}  for diastolic pressure.

Therefore, this thorough review of the existent literature on rescue robots reveals that the field of vital sign measurement is still in its nascent stages. While the measurement of temperature and respiratory rate can be accomplished with relative ease through the utilization of robotic systems, this level of ease is not consistently replicable when it comes to heart rate and, most notably, blood pressure. This is the context in which the present work is situated. It develops a mobile robotic system capable of automatically measuring pulse and blood pressure and tests the validity of its measurements and its performance in post-disaster environments.

The system comprises a quadruped robot equipped with an arm affixed to its back. This arm has, as a gripper, a soft device capable of determining the heart rate and the blood pressure. The operation of both the robot and the arm is remote, while the gripper's closure around the arm and the measurement process are fully automated.

The main contributions of this work relay on:
\begin{itemize}
    \item Firstly, the conceptualization and fabrication of a pneumatic soft gripper that exhibits the capacity to adapt to diverse human arms and interact with them in a risk-free manner.
    \item Secondly, the development of a portable pneumatic system that enables the inflation of the gripper is imperative. In the field of soft robotics, the issue of pneumatic portability stands as a particularly pressing challenge.
    \item Finally, the measurement of blood pressure, a metric that is rarely utilized in literature related to rescue operations, but has been demonstrated to offer substantial insights when combined with heart rate monitoring. This integration of data facilitates the development of more efficacious triage strategies, both for rescuers and intelligent systems.
\end{itemize}

The rest of the article is organized as follows. Initially, the various subsystems are presented, accompanied by a detailed exposition of the outcomes of these developments and the relegation of secondary design aspects to the supplementary material. Subsequently, the accuracy of the system is analyzed based on various statistical parameters, and its validity in SAR environments is demonstrated. Finally, the conclusions are drawn in Section 3.

\section{Results and Discussion}

\subsection{Overall System Architecture}

\begin{figure}[!ht]
  \includegraphics[width=\linewidth]{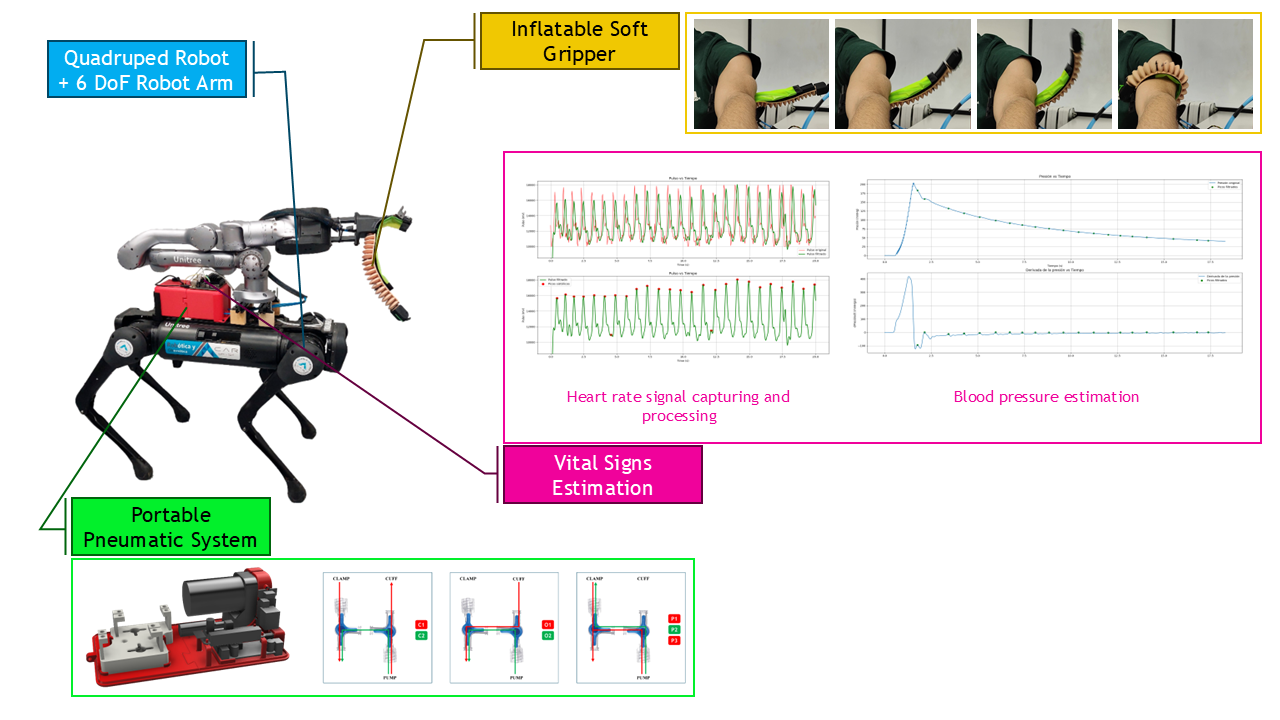}
  \caption{Diagram of the system architecture.}
  \label{fig:system-arch}
\end{figure}

The system, illustrated in Figure~\ref{fig:system-arch}, consists of four elements. The first one is a quadruped robot equipped with a 6-DoF arm. This robot is capable of navigating and moving in unstructured environments in search of victims. The second element of the system is the gripper of the manipulator. This element is composed of a soft actuator and a rigid clamp. The clamp closes around the patient's arm to obtain the heart rate curves. The third element comprises the air feeding to the gripper. The fourth subsystem is constituted by a Jetson Nano, which is responsible for the actuation of the valves in the pneumatic system and the processing of the information obtained from the gripper's sensors. This subsystem allows the estimation of the heart rate and blood pressure.

While the quadruped and the arm are teleoperated, the process of measuring vital signs is fully automated. Upon reception of the starting command, the Jetson Nano autonomously inflates the clamp and acquires data, facilitating the measurement of the heart curves of the victim. This data undergoes processing and, less than one second after, heart rate and arterial pressure are estimated.

\subsection{Mobile Robotic Platform}
\label{subsecc:robot}
The first subsystem comprises a quadrupedal robot equipped with an arm affixed to its back. The quadruped model, the Unitree Aliengo, has the capacity to navigate unstructured terrains, is equipped with a 2-hour battery life and possesses a payload capacity of up to \qty{10}{\kilo \gram}. This capacity facilitates the placement of the second element of the system, a Unitree Z1 manipulator, onto its dorsal surface. The arm weights \qty{4}{\kilo \gram} and possesses six degrees of freedom. The rest of the system's elements (the Jetson Nano, the pneumatic system, and the gripper) weigh \qty{3}{\kilo \gram}, leaving a \qty{1}{\kilo \gram} margin. Consequently, weight-carrying capabilities have been a critical factor in selecting the quadruped robot.

Although the subsystem is capable of autonomous navigation\cites{CruzUlloa2025}, for the purposes of this work, it has been teleoperated. This is because, in critical situations, full autonomy is not the best choice. It is preferable to have an operator in control when interacting with victims. Additionally, teleoperating the arm enables more precise placement of the gripper in the victim's hand. Immersive teleoperation interfaces based on virtual reality have been previously developed for non-robotic operators, i.e., professional rescuers, and allow precise control of the robot and arm\cites{CruzUlloa2024a}. 

\subsection{Gripper for Vital Signs Monitoring}

\begin{figure*}[!ht]
    \centering
    \begin{subfigure}[b]{0.45\textwidth}
        \centering
        \includegraphics[width=\linewidth]{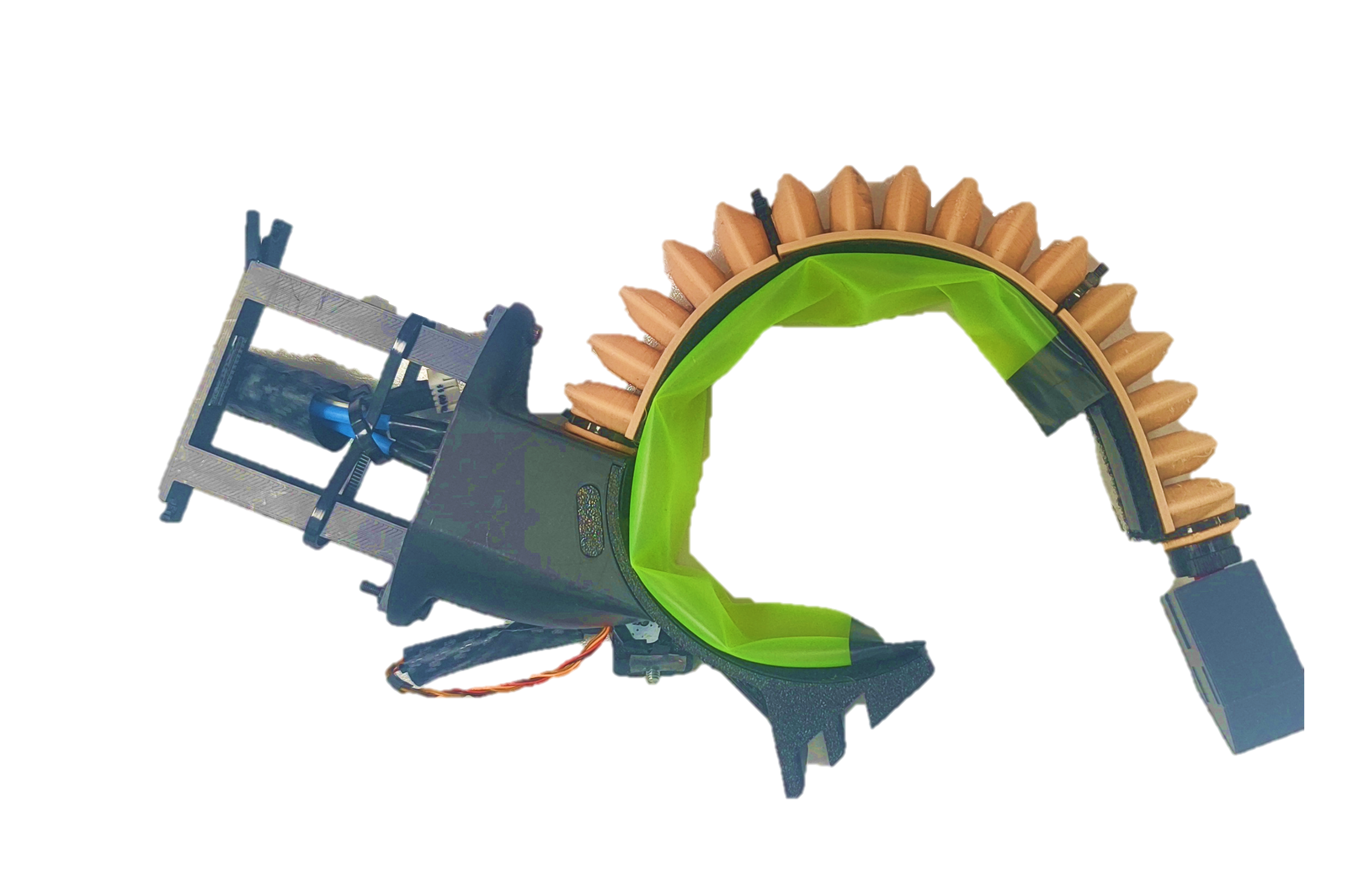}
        \caption{}
        \label{fig:gripper-picture}
    \end{subfigure}
    \hfill
    \begin{subfigure}[b]{0.45\textwidth}
        \centering
        \includegraphics[width=\linewidth]{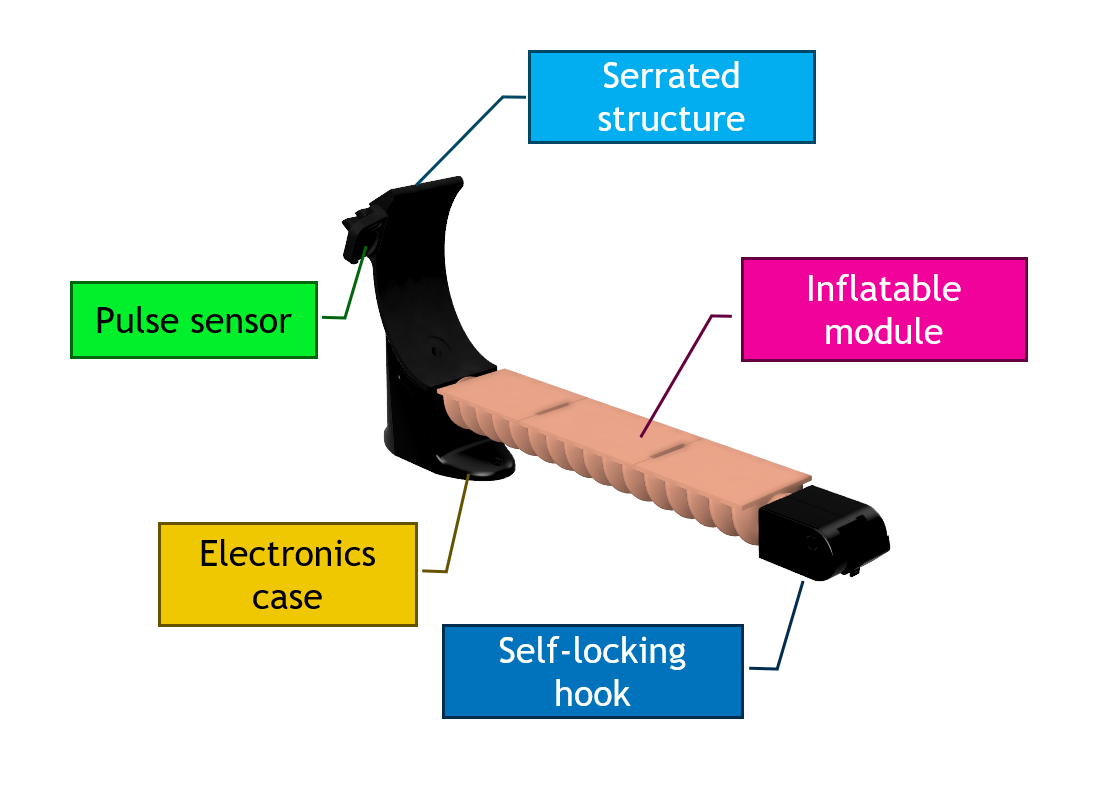}
        \caption{}
        \label{fig:gripper-clamp}
    \end{subfigure}
    \hfill
    \begin{subfigure}[b]{0.45\textwidth}
        \centering
        \includegraphics[width=\linewidth]{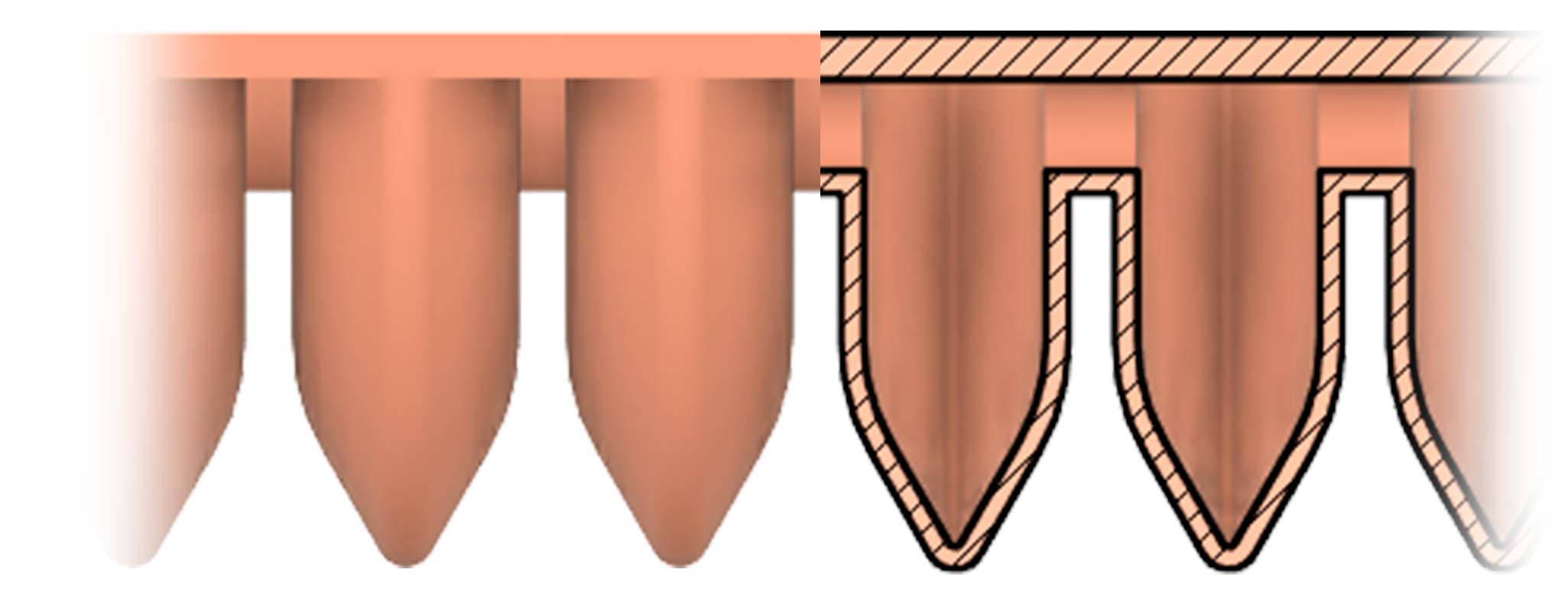}
        \caption{}
        \label{fig:gripper-inflatable}
    \end{subfigure}
    \hfill
    \begin{subfigure}[b]{0.45\textwidth}
        \centering
        \includegraphics[width=\linewidth]{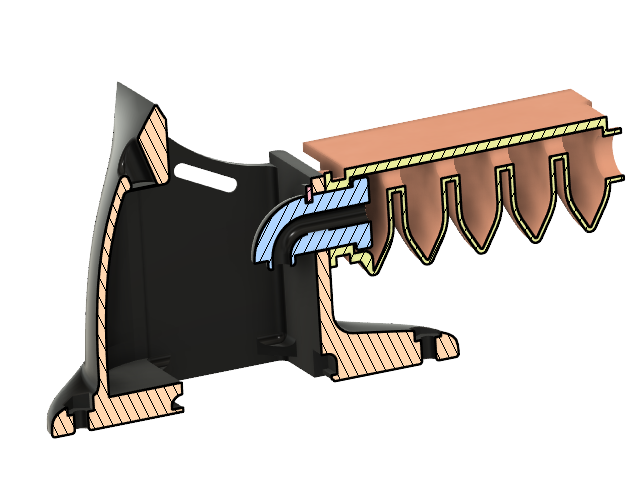}
        \caption{}
        \label{fig:gripper-conn}
    \end{subfigure}

    \caption{Design and components of the pneumatic gripper subsystem: 
        a) Picture of the subsystem.
        b) Components of the clamp. 
        c) Section of the inflatable module of the clamp. 
        d) Connection between the electronics case, where the pneumatic feed tube passes through, and the first segment of the inflatable module.}
    \label{fig:grippet}
\end{figure*}

The subsystem, a photograph of which can be seen in Figure~\ref{fig:gripper-picture}, is designed to measure two vital signs: heart rate and blood pressure. Both are measured in a similar way to digital blood pressure monitors. First, the victim's arm is wrapped. Then, the heart signal is stored and digitally processed to detect peaks. This allows the heart rate to be obtained, as well as the peaks linked to Korotkoff sounds. These sounds allow blood pressure to be determined.

Over non-invasive methodologies, tactile sensing has been identified as a preferred approach. This way is expected to demonstrate enhanced functionality in extreme conditions, such as low luminosity or the presence of fires.

To ensure the accuracy of heart rate measurements, it is imperative that the system maintain direct contact with the victim, applying constant pressure to ensure measurement stability without causing harm. The measurement of blood pressure, based on how a standard blood pressure cuff works, requires the device to wrap around the patient's arm, creating enough pressure to temporarily stop blood flow.

The primary design challenge was to ensure that the actuator can be positioned around the victim's arm when required, maintaining a secure grip during measurement, handling the pressure when the cuff inflates, and releasing safely when the measurement is done. Furthermore, the primary composition of the structure should consist of flexible materials manufactured using three-dimensional printing techniques. We chose this approach to reduce injury risk in the event of a collision.

The mechanism designed to fulfill these requirements is divided into two parts: a plastic cuff and a multi-component 3D-printed clamp, both equipped with independent pneumatic circuits, enabling each component to be inflated or deflated without interference from the other.

The cuff consists of a sealed plastic fabric measuring 5 $\times$ \qty{32.5}{\centi \metre} that is placed over the inflatable actuator. The function of this fabric is to inflate, as is the case with the cuff of a sphygmomanometer.

The clamp, on its side, consists of three components, all of them designed by 3D printing. The first one is the inflatable module, whose function is to envelop the arm. In order to guarantee that the module is firmly affixed to the arm, the second component of the clamp is a self-locking hook on one side and a serrated structure on the other side to which the hook is attached. The pulse sensor is also located at this end. In addition, the third component is the electronic case, which contains the devices responsible for converting the sensor signals from analog to digital. All of these components are depicted in Figure~\ref{fig:gripper-clamp}.

\subsubsection{Design and Manufacturing of the Inflatable Module}

The inflatable module is designed to envelop the subject's arm during the inflation process. The object under consideration consists of three segments that share the same design base, differing only in the joints at their ends. Figure~\ref{fig:gripper-inflatable} shows a section of the CAD design of the module, while Figure~\ref{fig:gripper-conn} shows the connection of the first segment to the end of the clamp. Details and images about the connection between segments are provided in Supplementary Note S1 and Supplementary Figures S1 and S2.

Each segment consists of a series of hollow, semicircular cavities that are connected to each other at the base. The sharp profile of these elements enhances their expansion capacity and reduces the internal overhang angle, thereby improving the quality of the finish during the 3D printing process. As the pressure within the cushions increases, they undergo expansion, resulting in the segment's curvature, with the curvature reaching up to \qty{360}{\degree}.
  
The module has been printed in TPU of 60A shore (Recreus FilaFlex TPU 60A) in a Bamu A1 mini printer. To date, the utilization of this shore has been limited, yet it possesses exceptional mechanical properties\cites{Gaafar2024, Lang2025}, since it poses a significantly greater challenge in terms of printing complexity. This discrepancy can be attributed, at least in part, to its higher level of elasticity exhibited.

However, this is an advantage in this application, since one of the major problems with printed pneumatic actuators is that they need to be designed with very thick walls to prevent air leaks, which makes them very rigid. Reducing the hardness up to 60A gives deformations analogous to those attained through silicone curing, but in a more repeteable way.

To ensure the impermeability of the module, it was imperative to correctly configure the print profile. A trial and error process was conducted, using the generic TPU profile as a starting point and making specific calibrations with the filament until an optimal configuration was achieved, as detailed in Supplementary Table S1.

The segments three have been printed with their bases resting on the printing surface. The hollow interior's printing has been optimized to eliminate the necessity for external supports. The application of supports has been restricted to the protrusions at the extremities, where tree-type supports have been utilized. Following the execution of an array of tests, it has been determined that this particular type of support was the optimal choice. The rationale for this determination is twofold: first, its diminished contact surface enables the efficient removal process without compromising the integrity of the component; secondly, it mitigates defects stemming from excessive adhesion, making the manufacturing process more repeatable.

\subsubsection{Sensorization and Control}
\label{subsub:sensors}
Two pressure sensors and a pulse sensor were used to measure vital signs. Both sensors return an analog signal that is transmitted to the Jetson Nano via ADS1115 ADC converters. A single pressure sensor was not enough as one sensor was used to measure blood pressure, and the other, to control the inflatable module in a closed loop. 

An optical sensor PPG-based has been utilized to measure the pulse. This non-invasive optical technique utilizes a sensor that emits infrared light in conjunction with a photodetector, which quantifies the amount of light reflected by the skin. The sensor functions with a supply voltage ($V_{DD}$) ranging from \qty{3}{\volt} to \qty{5.5}{\volt}, providing an analog output signal within the range of \qty{0.3}{\volt} to $V_{DD}$, with a center point at $\frac{V_{DD}}{2}$. The output signal is directly proportional to the detected blood flow.

The two pressure sensors are differential, model MPS20N0040D, and possess the capacity to measure pressures up to \qty{40}{\kilo \pascal} (approximately \qty{300}{\mmHg}). This range is sufficient for the measurement of blood pressure. These sensors provide a differential analog signal proportional to the measured pressure, which necessitates amplification. However, given the presence of a configurable amplifier in the ADC converters, the incorporation of an operational amplifier has become superfluous.

The ADS1115, is a 16-bit ADC converter equipped with an I2C interface. This device permits measurement in both differential and single-ended modes:
\begin{itemize}
    \item The pressure sensors are connected to an ADS1115 in differential mode, with an input range of $\pm$ \qty{0.512}{\volt}, thereby providing a resolution of \qty{15.6}{\micro \volt}. The ADDR pin is connected to GND in order to assign it the address 0x48.
    \item The temperature and pulse sensors are connected to the other ADS1115 in single-ended mode. The input range of the sensors is $\pm$ \qty{6.144}{\volt}, and the resolution is \qty{187.5}{\micro \volt}. The ADDR pin is connected to VCC in order to assign it the address 0x49.
\end{itemize}

\begin{figure}[!ht]
  \includegraphics[width=\linewidth]{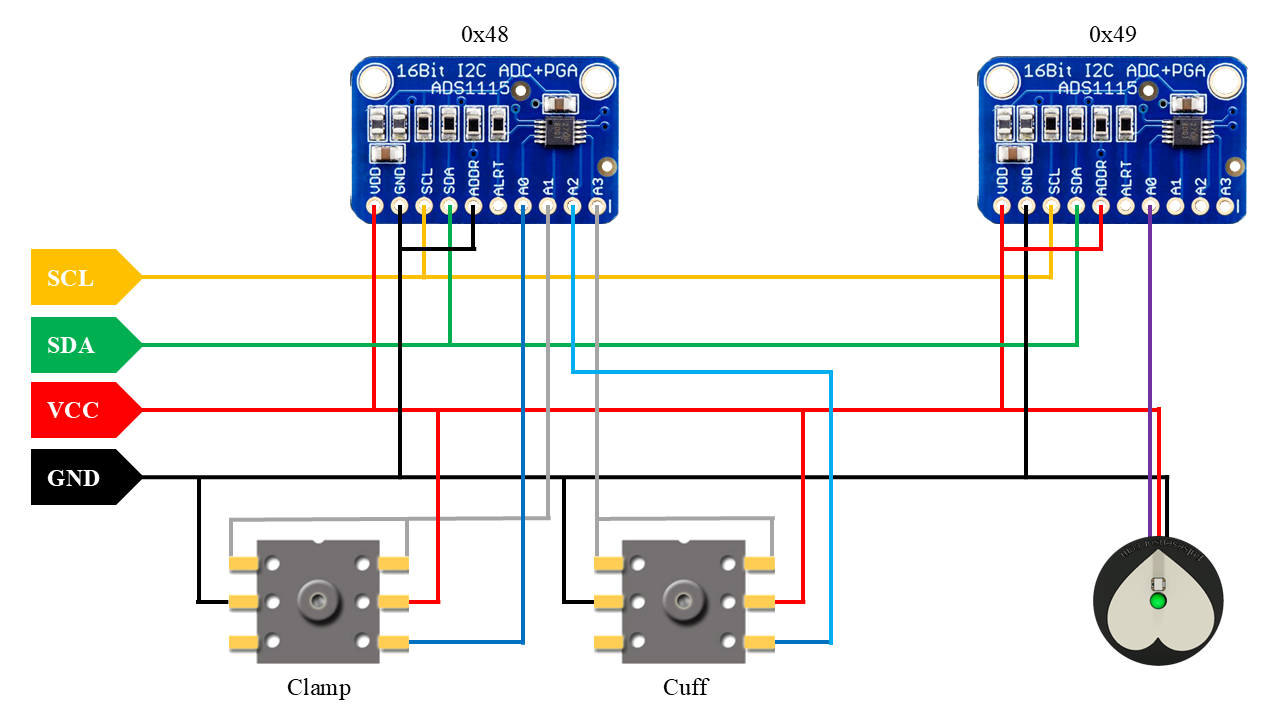}
  \caption{Schematic diagram of the system measurement elements.}
  \label{fig:sensors-circuit}
\end{figure}

A complete schematic diagram is depicted in Figure~\ref{fig:sensors-circuit}.

\subsection{Pneumatic Portability}

The issue of pneumatic system portability represents a significant challenge in the field of soft robotics research. A plethora of solutions have been attempted; however, only a limited number of them have demonstrated the capacity to generate adequate pressure to actuate devices such as the presented soft gripper and to be installed in a mobile robot.

Syringe-based solutions demonstrate a high-pressure regulation capacity; however, they are impractical when medium or large volumes of air are required\cites{Armiento2023}. Similarly, air cartridge-based systems are constrained by a limited number of actuations\cites{Drotman2021}. Electrode-based pumps are capable of high pressure but require a substantial power supply, which often results in a significant increase in weight\cites{Zhao2024}. Consequently, most portability-oriented solutions elect to utilize compact pumps and valves\cites{Cheung2024}.

In contrast to the approach outlined in reference\cites{Ahlquist2024}, the presented design design employs a pump capable of generating both vacuum and pressure. The use of a reservoir is not necessary, as the ripple can be effectively managed with the available air supply. This results in a reduction in the overall weight of the system.

\begin{figure}[!ht]
  \includegraphics[width=\linewidth]{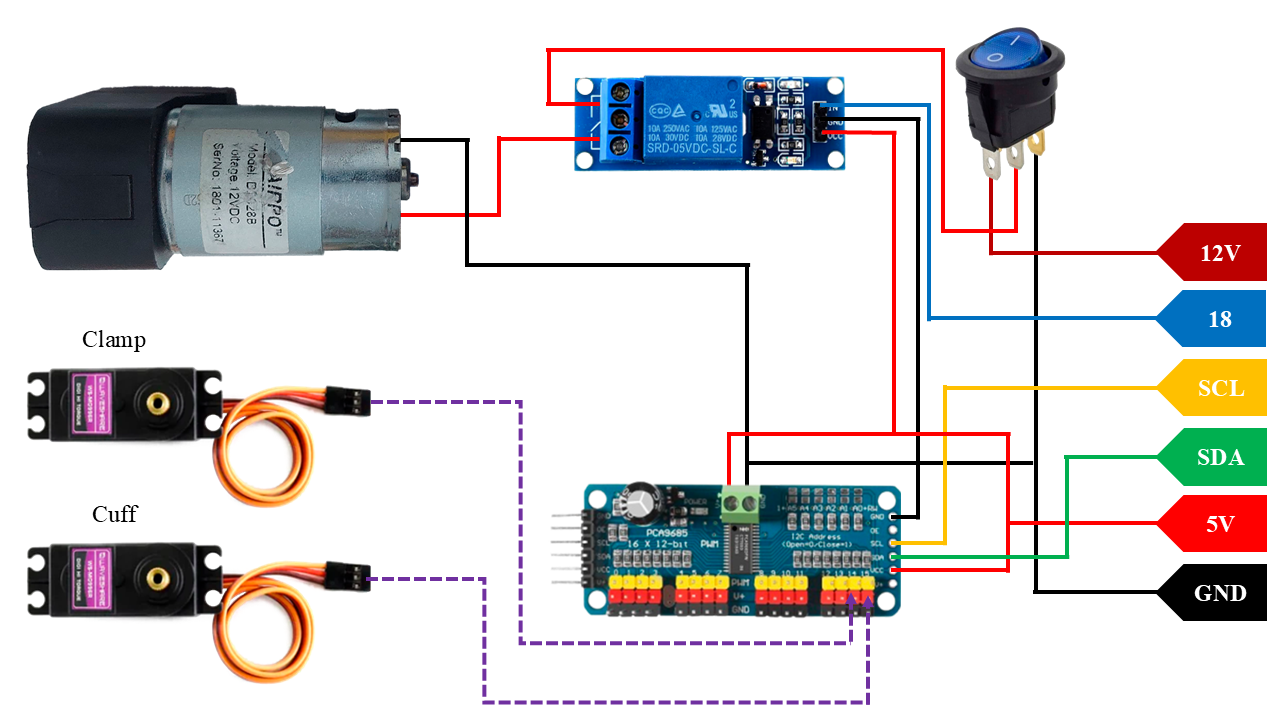}
  \caption{Schematic diagram of the pneumatic system.}
  \label{fig:pneumatic-circuit}
\end{figure}

The pneumatic system, illustrated in Figure~\ref{fig:pneumatic-circuit} is composed of a \qty{12}{\volt} pump, which is actuated by an electromechanical relay and two DC motors, the function of which is to regulate the flow of air into and out of the cuff and clamp from the pump. These devices operate at a voltage of \qty{5}{\volt}, necessitating the use of a DC-to-DC converter to facilitate power supply. A PCA9685 driver is used to control the motors. It receives the I2C signal from the Jetson Nano and transforms it into a PWM output. The system is also equipped with an emergency switch that enables its immediate deactivation in the event of overpressure, disconnections, or any other form of failure. All the elements are enclosed inside a 3D-printed box. Details regarding its geometry, the configuration of the elements internal to the structure to optimize spatial efficiency, and its position within the quadruped are provided in Supplementary Note S2 and in Supplementary Figures S3 and S4.

The pneumatic system is controlled by the central Jetson Nano. Its execution remains active as long as the user does not request the cancellation of the action. Furthermore, it incorporates an automatic timeout cancellation system, which halts the action if a value from the pressure sensor in charge of closing the loop is not received after \qty{500}{\milli \second} from the receipt of the last. A mechanism is also included that prevents execution from continuing in the event of a loss of connection with any of the other sensors, resuming only when that connection is re-established.

In the event of a system malfunction, the actuator is designed to return to a safe position. When this occurs, it is necessary to deactivate the pump and to completely release the pressure from both the actuator and the cuff. This precautionary measure is imperative to avert the potential risk of breakage in the event of pump failure.
 
In order to enhance the efficiency and stability of the system, sensor readings are managed using event-based programming. This mechanism enables the obstruction of pressure-dependent loops until a subsequent reading is obtained, thereby reducing the computational demand.

\begin{figure}[!ht]
     \centering
     \begin{subfigure}[b]{0.3\textwidth}
         \centering
         \includegraphics[width=\textwidth]{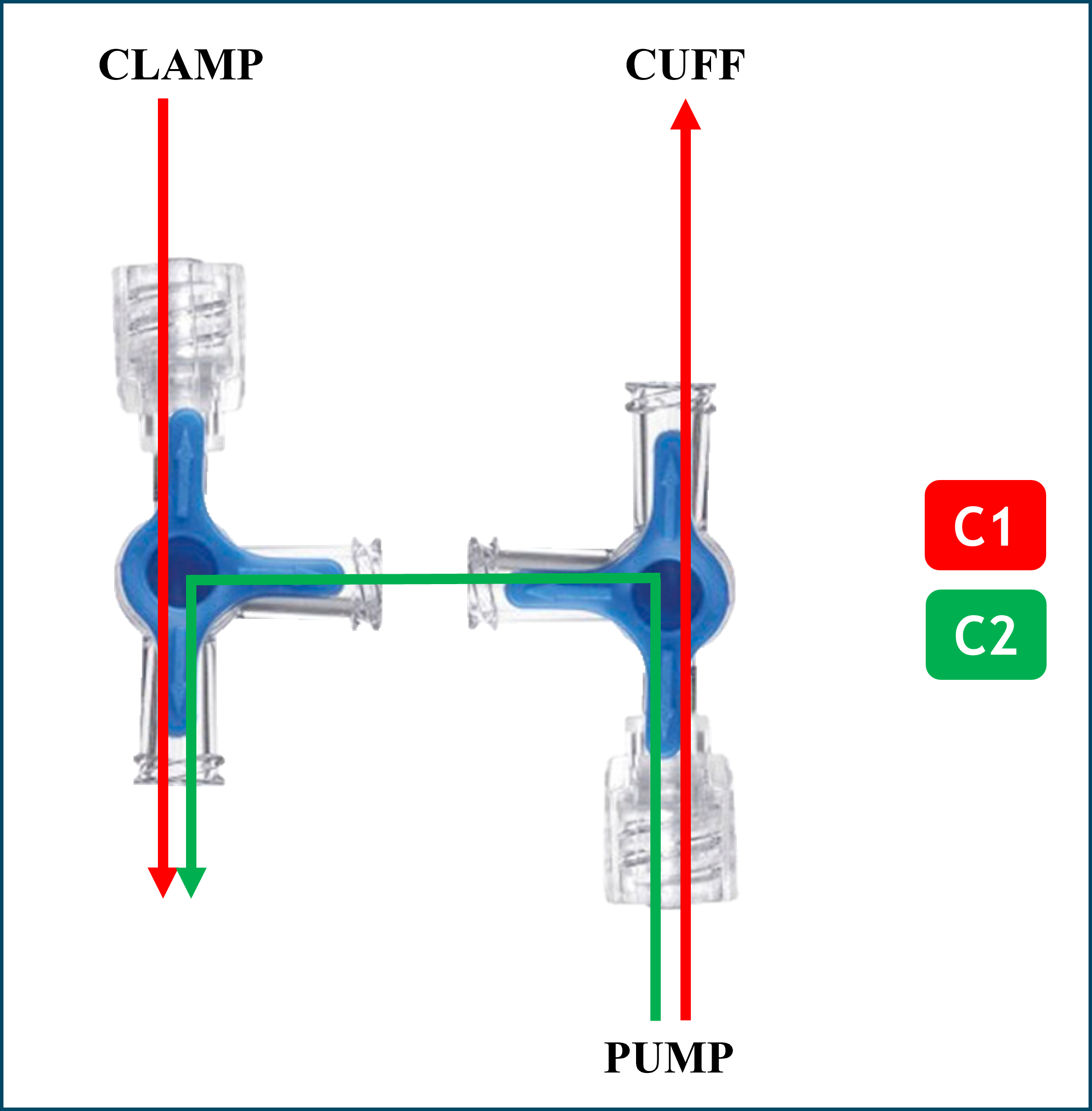}
         \caption{ }
         \label{fig:pneum-close}
     \end{subfigure}
     \hfill
     \begin{subfigure}[b]{0.3\textwidth}
         \centering
         \includegraphics[width=\textwidth]{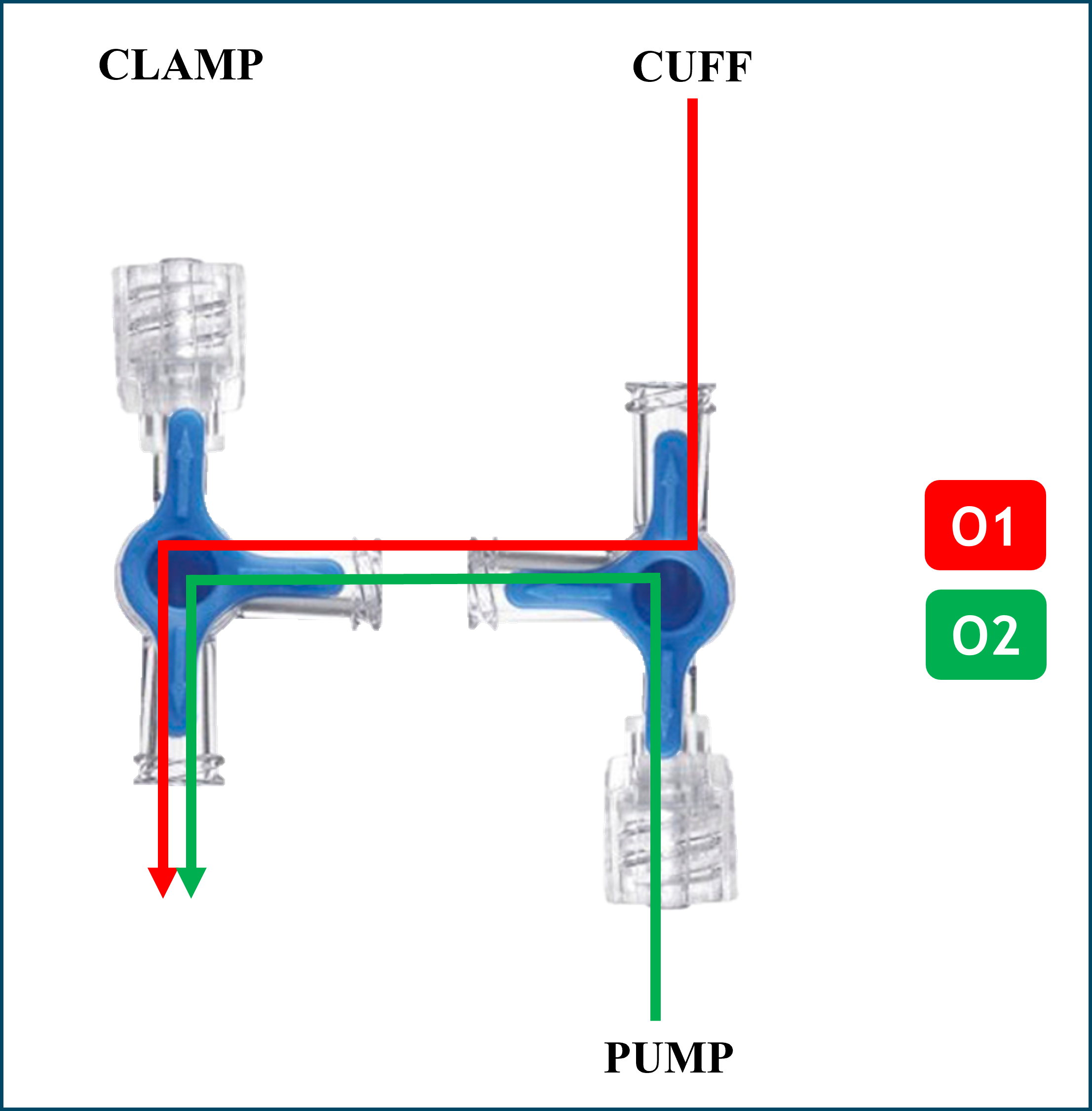}
         \caption{ }
         \label{fig:pneum-open}
     \end{subfigure}
     \hfill
     \begin{subfigure}[b]{0.3\textwidth}
         \centering
         \includegraphics[width=\textwidth]{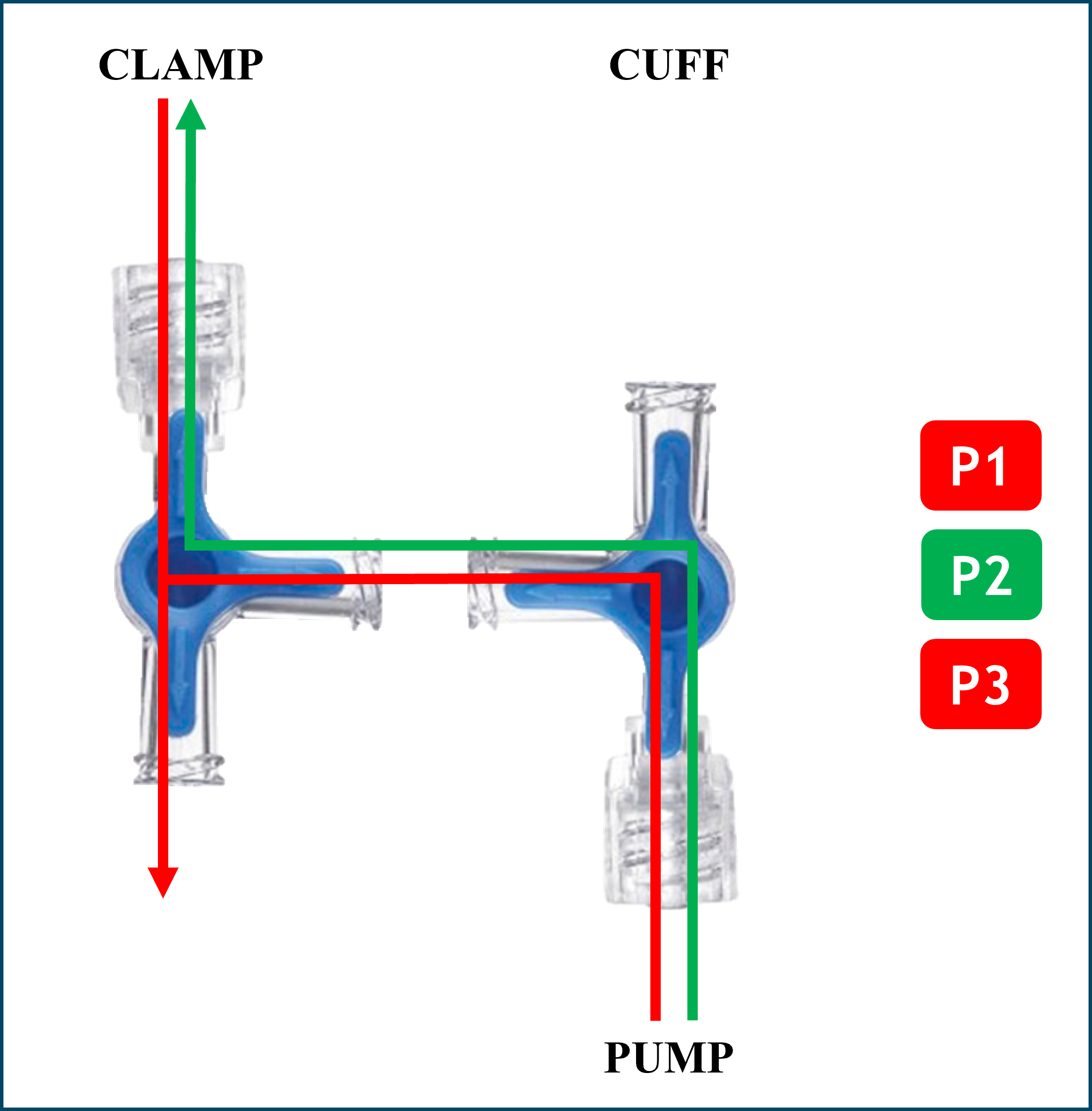}
         \caption{ }
         \label{fig:pneum-pressure}
     \end{subfigure}
     \hfill
     \begin{subfigure}[b]{\textwidth}
         \centering
         \includegraphics[width=\textwidth]{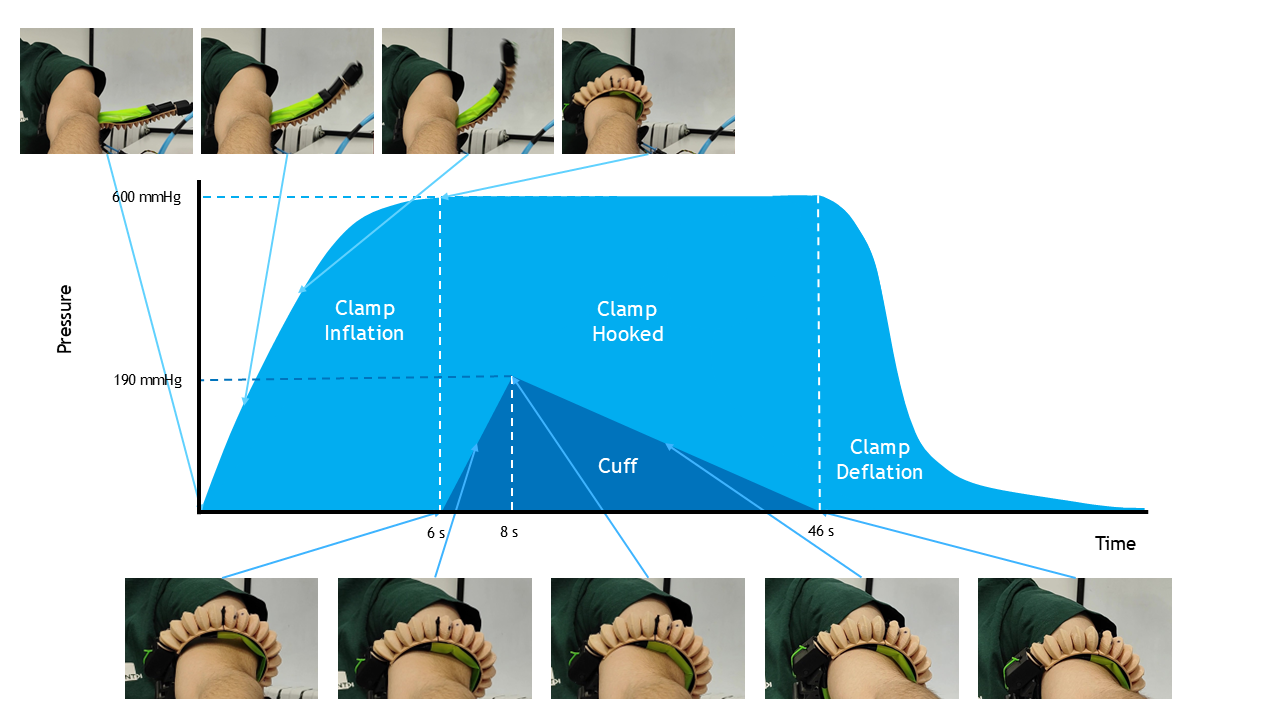}
         \caption{ }
         \label{fig:pneum-inflation}
     \end{subfigure}
    \caption{Functionality of the pneumatic system. a) Actuator closure. b) Actuator opening. c) Pressure measurement. d) Process of inflating and deflating the clamp and cuff. }
    \label{fig:pneum}
\end{figure}

The closing procedure of the actuator is depicted in Figure~\ref{fig:pneum-close}. Initially, the pressure sensor readings located within the clamp are initiated. Subsequently, the valves are configured to establish a connection between the air source and the clamp, thereby enabling the release of air that has been retained within the cuff (C1). This mechanism is designed to avert the potential malfunction of the actuator due to internal pressure buildup. Subsequently, the pump is activated. The process persists until the pressure in the clamp attains \qty{600}{\mmHg}, signifying that the actuator is fully closed and the self-locking hook piston is fully extended. Upon reaching this pressure threshold, the pump is deactivated, the pressure readings are halted, and the valves are adjusted to maintain the air within the system (C2). 

Figure~\ref{fig:pneum-open} illustrates the opening of the actuator. Upon receiving a command to open, the pressure sensor readings on the cuff are activated, and the valves are adjusted to permit the cuff to deflate (O1). Upon the decline in pressure to a virtually negligible level, the valves revert to their initial position, the sensor readings are deactivated, and a message confirming successful completion is disseminated (O2).

Finally, the blood pressure measurement process, which is described in Figure~\ref{fig:pneum-pressure}, is comprised of four phases. Initially, the inflatable cuff is completely deflated and the valves are adjusted to facilitate the release of air, while the corresponding sensor readings are initiated (P1). Subsequently, the valves are adjusted to establish a connection between the cuff and the pump (P2). The pump is activated until a pressure of \qty{190}{\mmHg} is attained. The subsequent controlled deflation phase, in where heartbeats are stored and processed to estimate pulse and blood pressure, is initiated. In this phase, the pressure is reduced at a steady rate of between 3 and \qty{5}{\mmHg} per second, thanks to the SL-type connector installed in the system. Once the pressure has dropped below the values at which it is reasonable to hear Kortkoff sounds, the residual air in the cuff is completely released (P3). 

An outline of the entire inflation and deflation process, with intermediate images of the condition of the cuff and clamp, is depicted in Figure~\ref{fig:pneum-inflation}.

\subsection{Signal Processing and Vital Signs Estimation}
\label{subsecc:signal-proc}

The process of vital signs estimation encompasses all the activities that allow the system to estimate the pulse and arterial pressures of the victim. As previously mentioned, the measurement and signal treatment processes are analogous to those of commercial digital sphygmomanometers. As delineated in Section~\ref{subsub:sensors}, the system comprises a pulse sensor and two pressure sensors, all of which are analogic. These devices are associated with two ADS1115 converters via the I2C bus. This configuration enables the Jetson Nano to receive the data and subsequently publish its readings via WiFi to the Jetson Nano.

The temporal framework within which each sensor type operates is delineated in Table~\ref{tab:sensors-freq}. It is has been decided that the interval between pulse sensor samples should not exceed 40 milliseconds. This requirement has been imposed to display photoplethysmograms in real time on web or mobile applications, giving the illusion of continuity. To achieve this, a frequency of 25 fps has been used, as is customary in the film industry. The pressure sensors, on their side, are monitored at a frequency four times higher. This is done in order to ensure that the peaks detected during the auscultation process are correctly identified, as they can be extremely low.

\begin{table}
 \caption{Sensor sampling frequencies}
 \label{tab:sensors-freq}
  \begin{tabular}[htbp]{@{}lrr@{}}
    \hline
    Sensor & Interval (\si{\milli\second}) & Frequency (\si{\hertz}) \\
    \hline
    Pressure (clamp)    & 10  & 100 \\
    Pressure (cuff)     & 10  & 100 \\
    Pulse               & 40  & 25  \\
    \hline
  \end{tabular}
\end{table}

\subsubsection{Heart Rate Processing and Estimation}
For the purpose of heart rate processing, the sampling rate is set at \qty{25}{\hertz}. This is managed by a callback function that determines the status of cardiac processing. During the active processing, the callback accumulates samples equivalent to \qty{200}{\milli \second} and processes them. The selection of this interval was driven by the objective of mitigating the computational demands, thereby introducing a negligible delay that is imperceptible to the user.

\begin{figure*}[!ht]
    \centering
    \begin{subfigure}[b]{\textwidth}
        \centering
        \includegraphics[width=\linewidth]{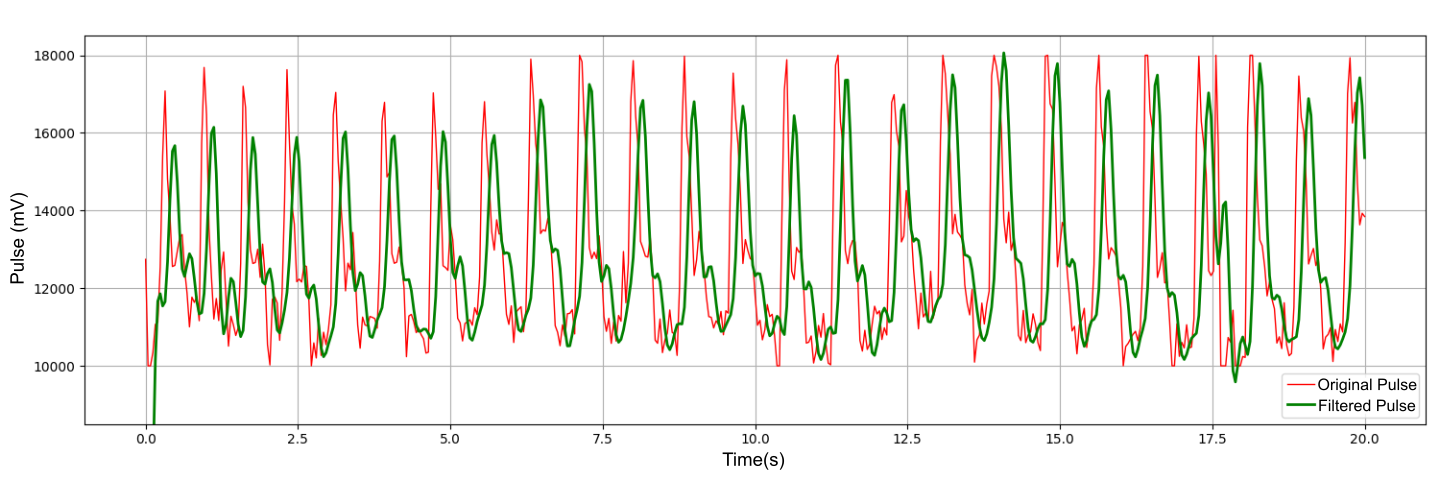}
        \caption{~}
        \label{fig:signal-pulse}
    \end{subfigure}
    \hfill
    \begin{subfigure}[b]{\textwidth}
        \centering
        \includegraphics[width=\linewidth]{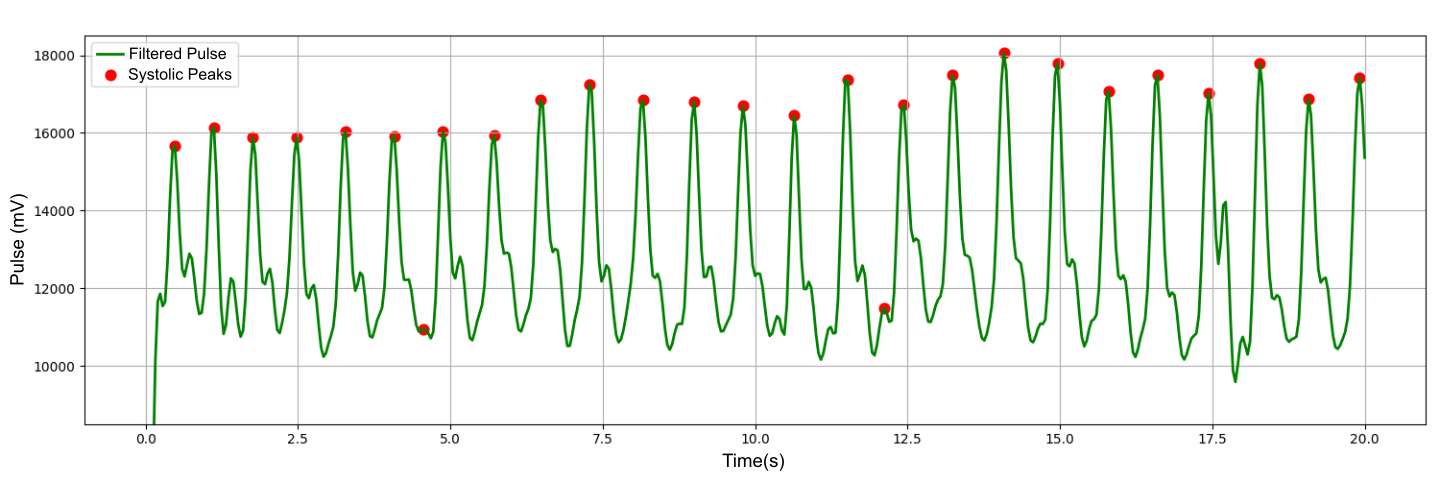}
        \caption{~}
        \label{fig:signal-peaks}
    \end{subfigure}
    \hfill
    \begin{subfigure}[b]{\textwidth}
        \centering
        \includegraphics[width=\linewidth]{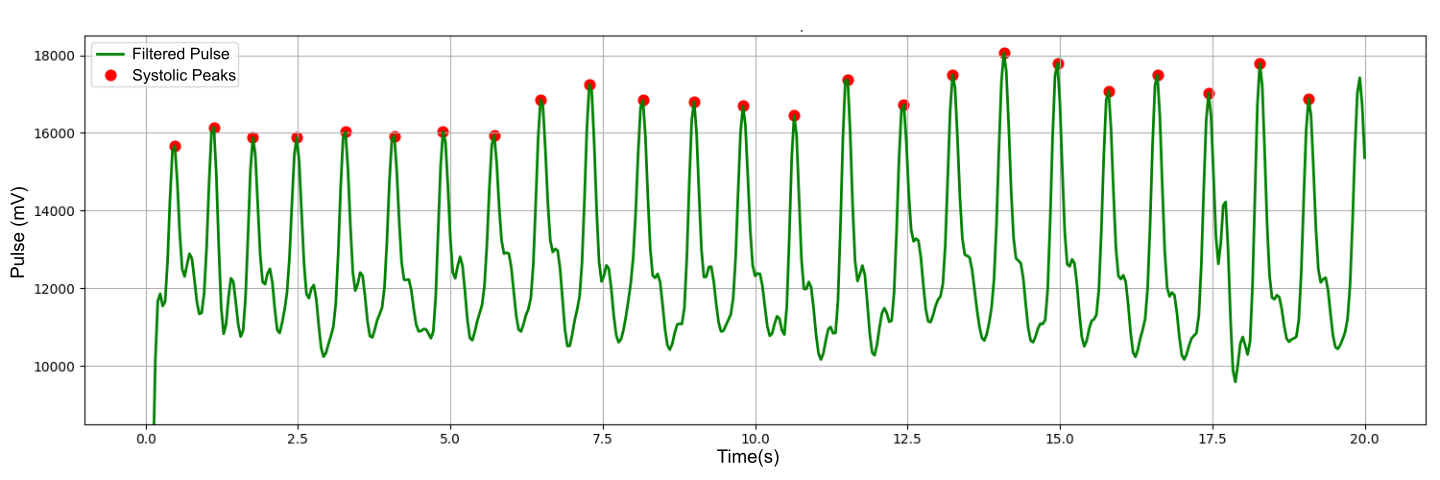}
        \caption{~}
        \label{fig:signal-peaks2}
    \end{subfigure}

    \caption{Pulse signal processing stages: 
        a) Low-pass filtering at \qty{3.5}{\hertz} of the complete pulse signal.
        b) Systolic peaks detected in the first phase. 
        c) Systolic peaks detected after applying the adaptive filter based on prominence.}
    \label{fig:signal-combined}
\end{figure*}

Initially, a fourth-order Butterworth IIR low-pass filter is employed, as presented in Figure~\ref{fig:signal-pulse}, with a cutoff frequency of \qty{3.5}{\hertz}. This method effectively eliminates high-frequency noise without introducing artificial oscillations. The application of the filter is contingent upon the preservation of the internal state, $z_{i}$, which is necessary to maintain the continuity of filtering between signal fragments.

Subsequently, the parameters of interest, including PPM, IBI, frequency, SDNN, and RMSSD, are extracted. The definitions of all these variables can be found in Supplementary Note S3. In both processes, if processing is completed or an error is detected, a value of -1 is written in the corresponding field as an indicative of a malfunction event.

This second step of extracting cardiac information is entirely based on peak detection. However, this is not trivial, as the signal can be affected by artifacts generated by patient movement, and its morphology depends on blood flow in the measurement area, which introduces high variability between subjects and between different possible sensor locations.

Initially, all peaks that meet a minimum separation between them of \qty{0.3}{\second}, equivalent to a heart rate of \qty{200}{\bpm}, are selected. This condition enables the elimination of high-frequency oscillations that do not correspond to actual heartbeats. Figure~\ref{fig:signal-peaks} illustrates this step.

However, it should be noted that these peaks may include false positives, thus necessitating the implementation of an adaptive filter based on prominence. Given the approximately periodic nature of the pulse wave and its relatively constant amplitude, peaks that exhibit a prominence between 40\% below and 60\% above the average prominence of the peaks recorded in the previous 20 seconds are considered valid, as shown in Figure~\ref{fig:signal-peaks2}, in which peaks like the one occurring at time $t = $ \qty{12}{\second} is not longer detected as a main peak.

\subsubsection{Blood Pressure Processing}

\begin{figure}[!ht]
  \includegraphics[width=\linewidth]{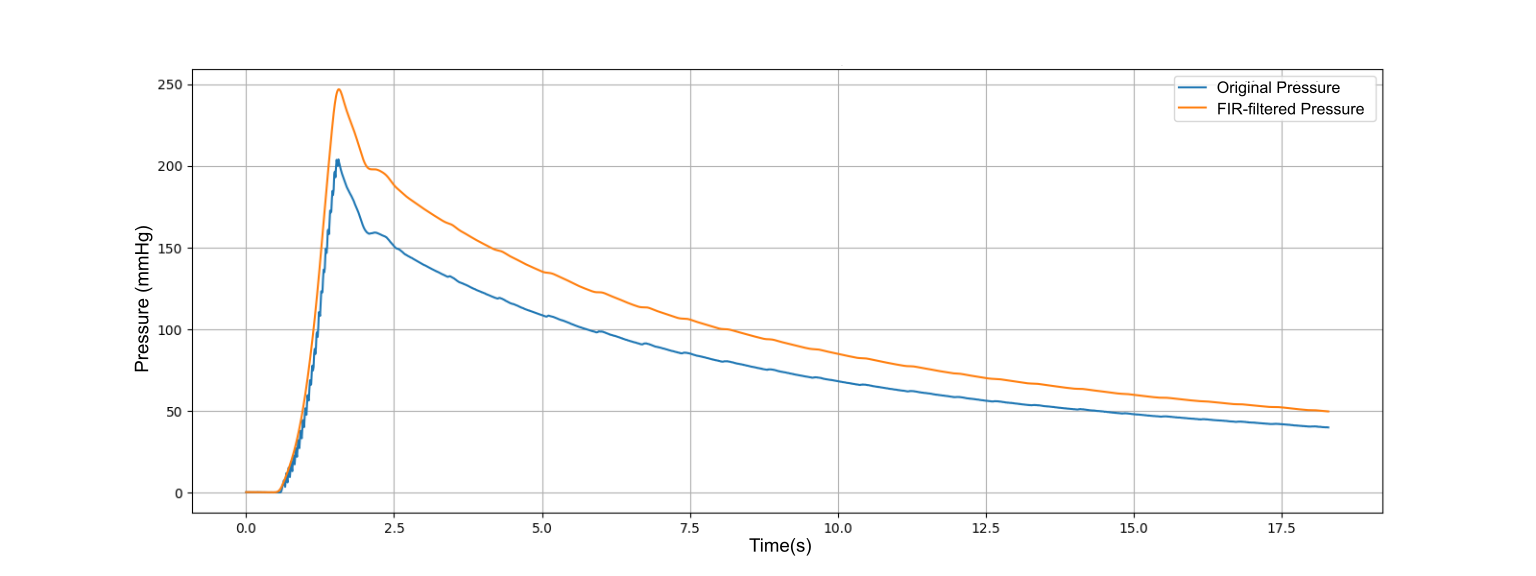}
  \caption{Comparison between the original pressure signal and its filtered version.}
  \label{fig:press-filter}
\end{figure}

The calculation of blood pressure is achieved through the utilization of pressure sensors operating at a frequency of \qty{100}{\hertz}. The initial step in this process entails the extraction of systolic and diastolic pressure values from a compendium of pressure readings. In order to execute this process, the signal undergoes a preliminary filtration stage that employs a bandpass filter with a frequency range spanning from 0.5 to \qty{4}{\hertz}. The filter operates on both forward and backward data, thereby eradicating the phase shift that is introduced by the filtering process. The result of it can be observed in Figure~\ref{fig:press-filter}.

The employment of an FIR filter over an IIR filter is predicated on the superiority of the former in preserving the waveform, a critical consideration in this particular scenario. A minor alteration in the position of the morphological peaks has the potential to exert a substantial influence on the calculation of blood pressure.

\begin{figure}[!ht]
  \includegraphics[width=\linewidth]{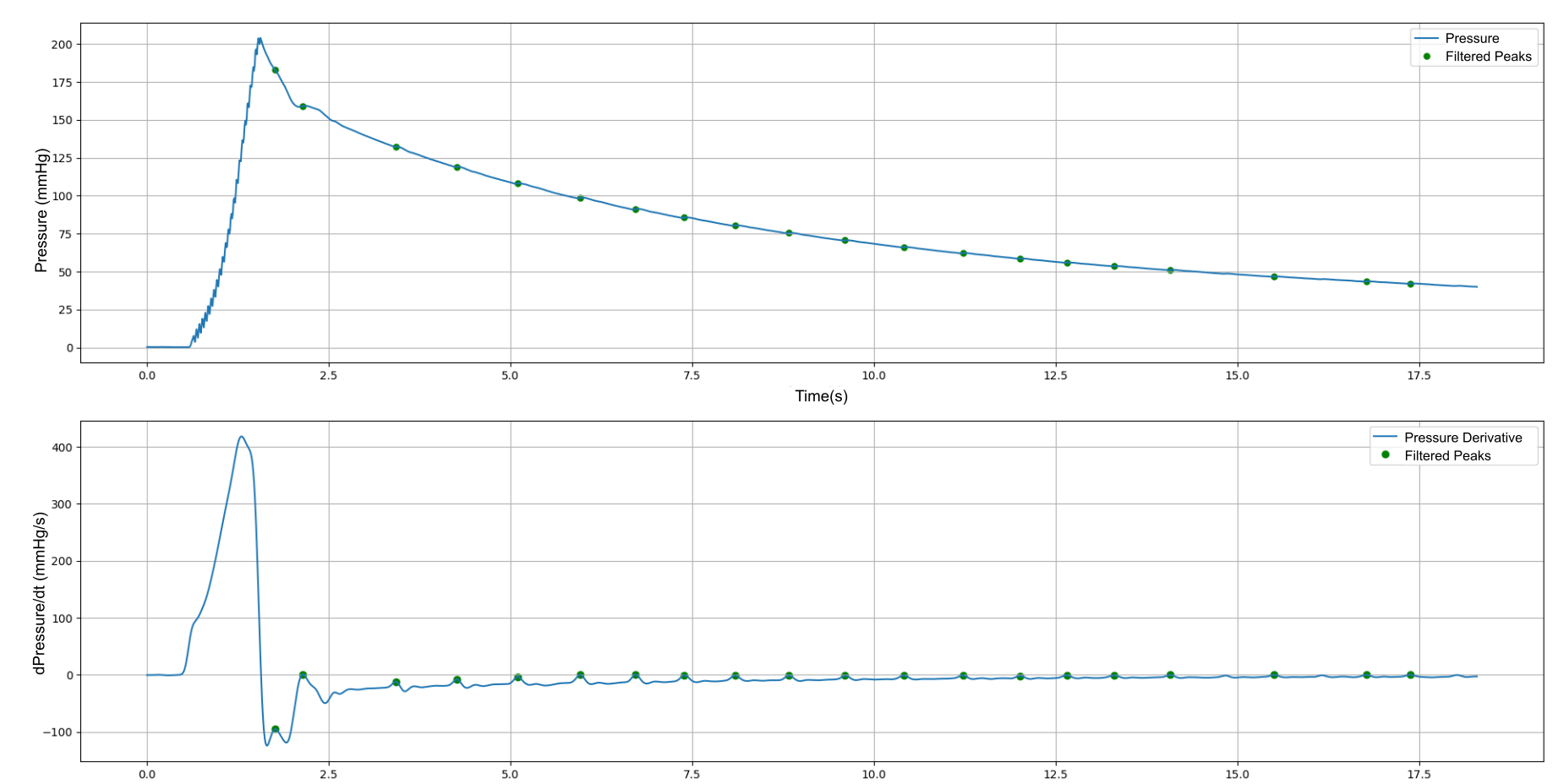}
  \caption{Peaks detected after morphological filtering.}
  \label{fig:press-filter-morph}
\end{figure}

Subsequently, the derivative of the filtered pressure signal is obtained for later use in identifying the pulses caused by the subject's heartbeats. Over this signal, a new morphological filter is applied to this derived signal to extract the peaks generated by the heartbeats. To accomplish this objective, peaks that satisfy specific height and prominence criteria are searched. In order to accurately assess the prominence of a peak, it is imperative to specify a window width that is equivalent to twice the maximum number of samples anticipated between pulses. This ensures an adequate local evaluation, with the minimum heart rate of \qty{40}{\bpm} being the baseline. Furthermore, peaks detected during the cuff inflation phase are removed. Results of the process are displayed in Figure~\ref{fig:press-filter-morph}.

\begin{figure}[!ht]
  \includegraphics[width=0.6\linewidth]{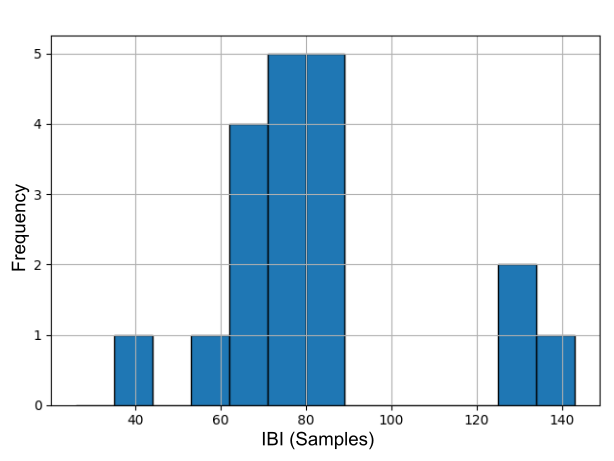}
  \caption{Histogram of the distances (in number of samples) between consecutive peaks.}
  \label{fig:press-hist}
\end{figure}

In a subsequent step, a secondary filter is applied, based on the distance between the peaks. The underlying assumption is that the pulses must be grouped together and exhibit a certain temporal regularity. The purpose of this filter is to discard peaks that do not meet these conditions. The implementation of the filter is initiated by the calculation of a histogram of the distances between consecutive peaks, displayed in Figure~\ref{fig:press-hist}. This calculation is performed using intervals that encompass the range of heart rates between 40 and \qty{230}{\bpm}, with a step size of 90 milliseconds.

The utilization of this histogram is predicated on the objective of ascertaining the most representative distance range. To accomplish this objective, a maximum of five non-zero intervals, each containing the highest total number of pulses, are selected. The interval between the start of the first interval and the end of the last interval is then defined as the valid range. Subsequent to determining this range, the peaks are filtered, thereby eliminating those whose distance from the next peak is outside this interval. Among the resulting groups, it is imperative to retain solely the longest group of consecutive peaks.

\begin{figure}[!ht]
  \includegraphics[width=\linewidth]{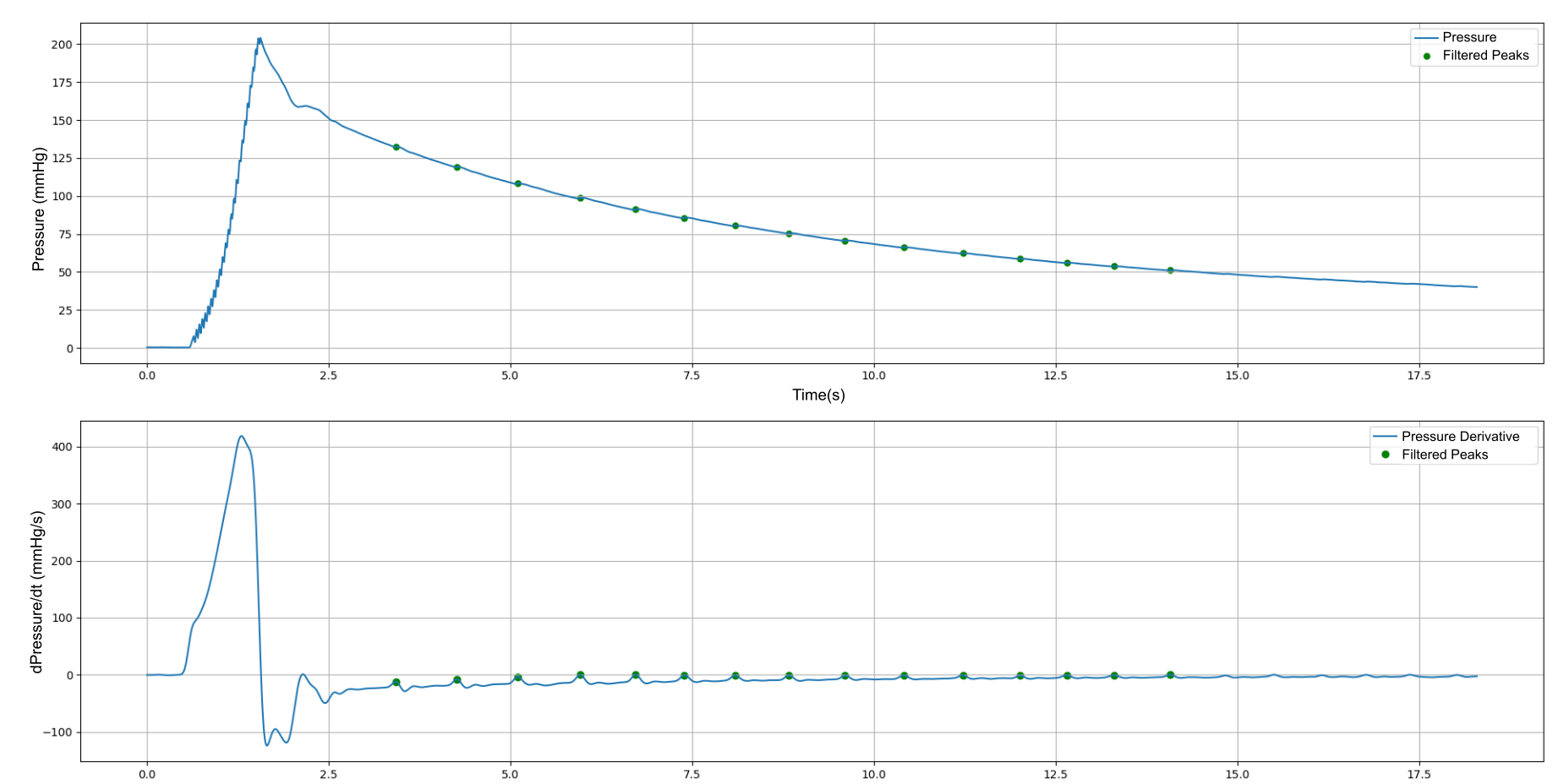}
  \caption{Peaks detected after distance filtering.}
  \label{fig:press-filter-dist}
\end{figure}

In order to prevent isolated peaks generated by noise from interrupting a valid group, the continuity of the group is maintained if the distance between a peak and the one following the next one (i.e., the sum of two consecutive distances) remains within the allowed distance range. The result of this filtering is depicted in Figure~\ref{fig:press-filter-dist}

The peaks detected following this final filtration are regarded as representative of the heartbeats, and from these, the systolic and diastolic pressures are calculated. The majority of commercially available digital sphygmomanometers determine these pressures by first identifying the mean arterial pressure (MAP) and then estimating systolic and diastolic pressures using empirical proportions of the oscillometric profile.

\begin{figure}[!ht]
  \includegraphics[width=0.6\linewidth]{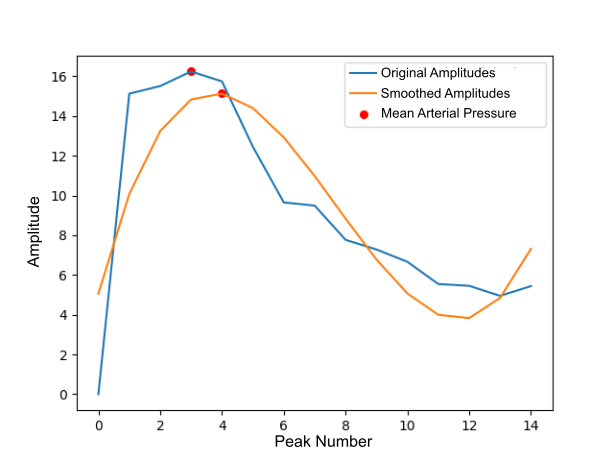}
  \caption{Original and smoothed amplitudes and location of MAP.}
  \label{fig:press-map}
\end{figure}

The MAP value is determined by first calculating the amplitudes of the valid peaks. In order to prevent incorrect amplitudes caused by noise from affecting the MAP determination, the amplitude signal is smoothed by applying a third-degree polynomial fit. This is illustrated in Figure~\ref{fig:press-map}. This technique enables the elimination of irregularities without compromising the overall structure of the oscillometric envelope. The peak corresponding to the maximum amplitude in the smoothed signal is identified as the MAP peak.

\begin{figure}[!ht]
  \includegraphics[width=\linewidth]{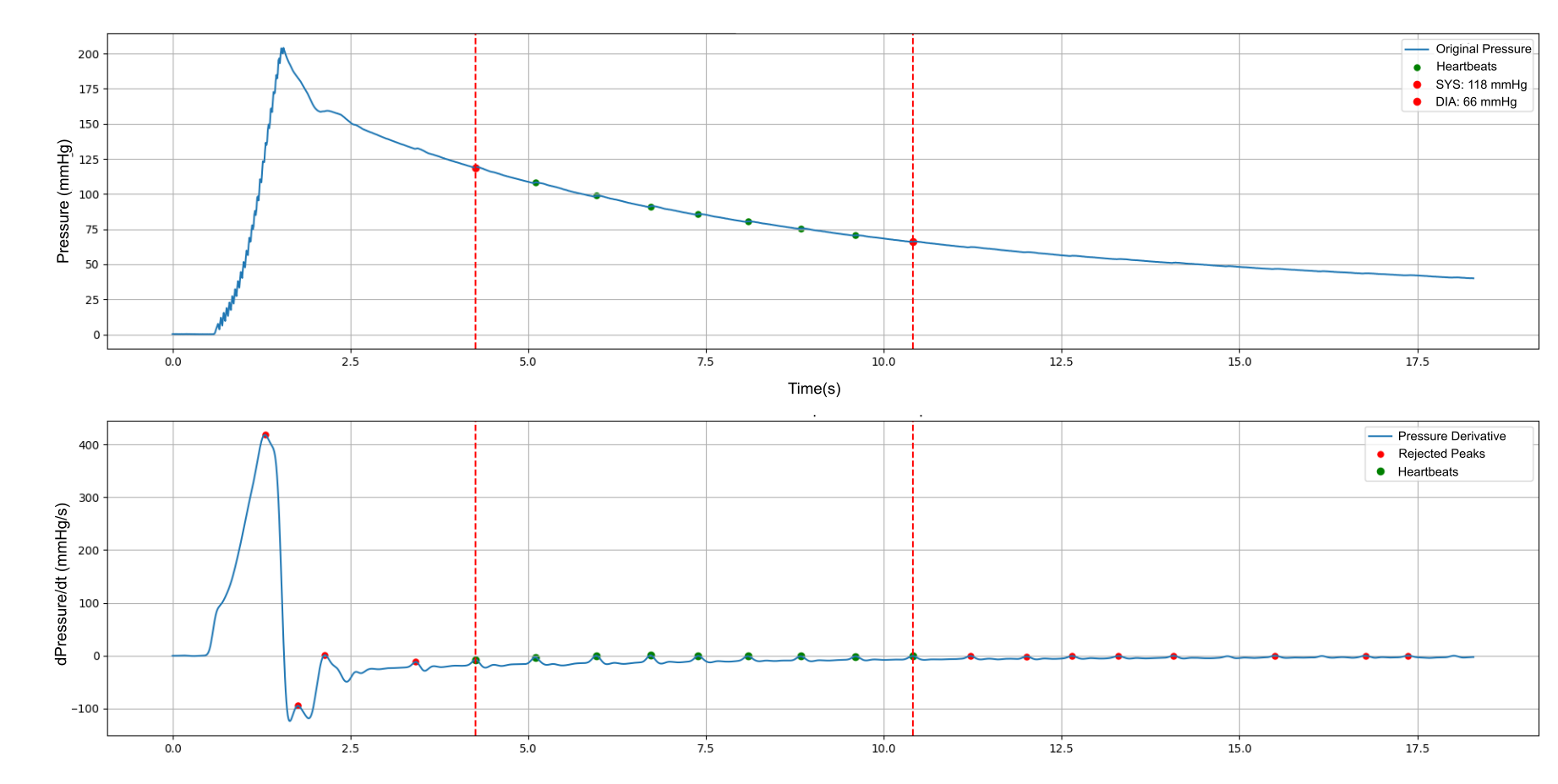}
  \caption{Graphs of the pressure signal and its derivative, showing the blood pressure values obtained and all the peaks analyzed.}
  \label{fig:press-sbp-dbp}
\end{figure}

Following the identification of the MAP, the systolic pressure is calculated as the closest peak prior to the MAP whose amplitude, in conjunction with that of the peak preceding it, does not exceed 88\% of the MAP amplitude. Similarly, diastolic pressure is determined as the first peak after the MAP whose amplitude, together with that of the next peak, does not exceed 42\% of that amplitude. This double-checking process, which also incorporates the adjacent peak, serves to mitigate the effect of erratic peaks resulting from noise. An example of this final output is presented in Figure~\ref{fig:press-sbp-dbp}.

Subsequent to the calculation of systolic and diastolic pressures, their values are examined to ascertain that they are within physiologically plausible ranges --intervals $[70, 240]$ \unit{\bpm}  and $[40, 140]$ \unit{\bpm} for systolic and diastolic pressures, respectively--. In the event that this is the case, the values in question are returned in conjunction with the estimated heart rate, which is calculated from the mean value of the intervals between the peaks of systolic and diastolic pressures.

If is not the case, a -1 value is generated to indicate an error and request that the robotic system initiate the measurement process again. In the event that the process encounters failure on three consecutive occasions, further measurements are suspended, and a notification is disseminated to the rescue personnel, indicating the potentiality of the victim's demise.

\subsection{Performance Analysis}

This section presents the outcomes of the validation process conducted on the pulse and blood pressure measurement subsystem. The subsequent section is concerned with the validation of the robotic system in its entirety. This distinction was made because this subsystem was considered to be the central one of all those that constitute the architecture presented here.

\subsubsection{General Error Analysis}
The experiments were conducted under the conditions delineated in Section 4. Twenty volunteers, aged between 21 and 58, were selected for the tests. The study identified two limitations. Firstly, the sample size is limited. The majority of validation standards for blood pressure measurement devices, for instance ISO 81060-2:2018, stipulate a sample size of $N = 85$ (and a minimum of 255 valid pairs)\cites{ISO81060-2:2018}. However, certain standards, including that of the European Society of Hypertension (ESH-IP2), decrease this requirement to $N = 33$\cites{Stabouli2022}. The second limitation is that volunteers with extreme blood pressure ranges were not actively sought, as demonstrated in the Supplementary Figures S7 and S8.

It is important to note, however, that the purpose of the present study is to serve as an exploratory test of the system's performance. Moreover, the primary function of the system is to estimate pulse and blood pressure, thereby assisting rescuers in their decision-making and triage, rather than providing an entirely precise estimation.

Table~\ref{tab:perf-err} shows the values for Mean Absolute Error (MAE), Root Mean Square Error (RMSE), Median Absolute Error (MedAE), and percentage error. The use of MedAE is further justified by the low number of samples, as it is more robust against outliers.

\begin{table}[!ht]
    \caption{Errors in pulse and blood pressure measurement}
    \label{tab:perf-err}
    \begin{tabular}{@{}lrrrr@{}}
        \hline
        ~ & MAE & RMSE & MedAE & \% Error (MAE) \\ 
        \hline
        Pulse & \qty{9.63}{\bpm} & \qty{14.93}{\bpm} & \qty{7.00}{\bpm} & 13.18 \% \\
        Pulse (w/o outlier) & \qty{7.20}{\bpm} & \qty{9.20}{\bpm} & \qty{6.00}{\bpm} & 9.5 \% \\ 
        Systolic Pressure & \qty{12.06}{\mmHg} & \qty{16.00}{\mmHg} & \qty{9.67}{\mmHg} & 10.65 \% \\ 
        Diastolic Pressure & \qty{9.50}{\mmHg} & \qty{12.22}{\mmHg} & \qty{8.00}{\mmHg} & 13.01 \% \\ 
        \hline
    \end{tabular}
\end{table}

With regard to heart rate, a moderate average error is evident, and as indicated by the MedAE value, the majority of measurements differ from the actual pulse value by less than \qty{7}{\bpm} This suggests that the system is capable of providing a reasonably accurate estimate of the victim's pulse. The elevated RMSE value signifies a substantial dispersion of outcomes. A thorough examination of the results, presented in Supplementary Table S2, reveals the presence of an outlier in element N=14, which significantly distorts the sample. In the absence of this adjustment, errors are much lower, indicating a high degree of accuracy in the system's estimation of heart rate.

Concerning pressures, the device's capacity to estimate them is evident, despite minor inaccuracies. The system demonstrates an ability to estimate pressures with errors of approximately 10\%, with half of the measurements falling below \qty{10}{\mmHg} for systolic pressure and \qty{8}{\mmHg} for diastolic pressure.

Supplementary Table S3 presents the errors corresponding to deciles 1, 5 (although this coincides with the MedAE values), 7, and 9, while Supplementary Figures S10-S12 illustrate their statistical distribution. The findings from both sets of results indicate that the majority of the measurements yield highly precise outcomes, with distributions exhibiting a pronounced left skew. This suggests that, despite its imperfections, which slightly exceed those found in the literature on medical devices, the system possesses a reasonable predictive capacity.

\subsubsection{Systemic Error Analysis}
To assess the potential systemic error of the developed meter, an agreement analysis presented in Table~\ref{tab:perf-bias}. The formulas for calculating its metrics, as well as the limits indicated in the Bland-Altman diagram of Figure~\ref{fig:bland-altman}, are detailed in Supplementary Note S4. Table~\ref{tab:perf-bias-wo-out} reflects the results of the same analysis after removing the outlier of the element N=14.

\begin{table}[!ht]
    \caption{Agreement analysis between the robotic system and the reference sphygmomanometer.}
    \label{tab:perf-bias}
    \begin{tabular}{lcccc}
    \hline
    Parameter & Bias & SD & LoA low & LoA high \\
    \hline
    Pulse & $-\qty{6.68}{\bpm}$ & \qty{13.69}{\bpm} & $-\qty{33.52}{\bpm}$ & \qty{20.16}{\bpm} \\
    Systolic Pressure & $-\qty{6.07}{\mmHg}$ & \qty{15.19}{\mmHg} & $-\qty{35.84}{\mmHg}$ & \qty{23.70}{\mmHg} \\
    Diastolic Pressure & $-\qty{6.92}{\mmHg}$ & \qty{10.33}{\mmHg} & $-\qty{27.16}{\mmHg}$ & \qty{13.33}{\mmHg} \\
    \hline
    \end{tabular}
\end{table}

\begin{table}[!ht]
    \caption{Agreement analysis between the robotic system and the reference sphygmomanometer after removing the outlier of the element N=14.}
    \label{tab:perf-bias-wo-out}
    \begin{tabular}{lcccc}
    \hline
    Parameter & Bias & SD & LoA low & LoA high \\
    \hline
    Pulse & -\qty{4.21}{\bpm} & \qty{7.37}{\bpm} & -\qty{18.65}{\bpm} & \qty{10.22}{\bpm} \\
    Systolic Pressure & -\qty{4.94}{\mmHg} & \qty{11.71}{\mmHg} & -\qty{27.88}{\mmHg} & \qty{17.99}{\mmHg} \\
    Diastolic Pressure & -\qty{6.25}{\mmHg} & \qty{9.46}{\mmHg} & -\qty{24.79}{\mmHg} & \qty{12.29}{\mmHg} \\
    \hline
    \end{tabular}
\end{table}

\begin{figure}
     \centering
     \begin{subfigure}[b]{0.45\textwidth}
         \centering
         \includegraphics[width=\textwidth]{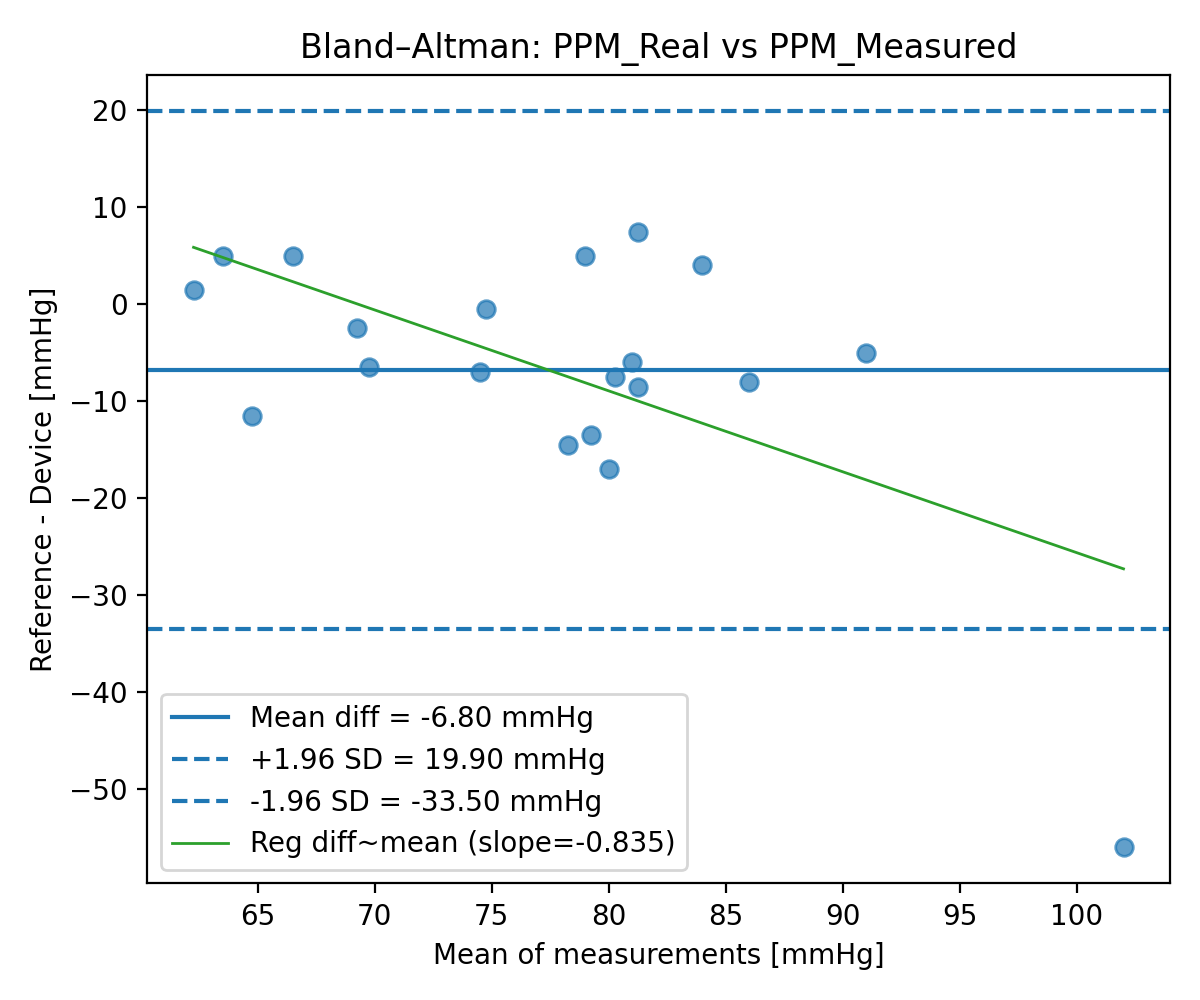}
         \caption{ }
         \label{fig:bland-altman-pulse}
     \end{subfigure}
     \hfill
     \begin{subfigure}[b]{0.45\textwidth}
         \centering
         \includegraphics[width=\textwidth]{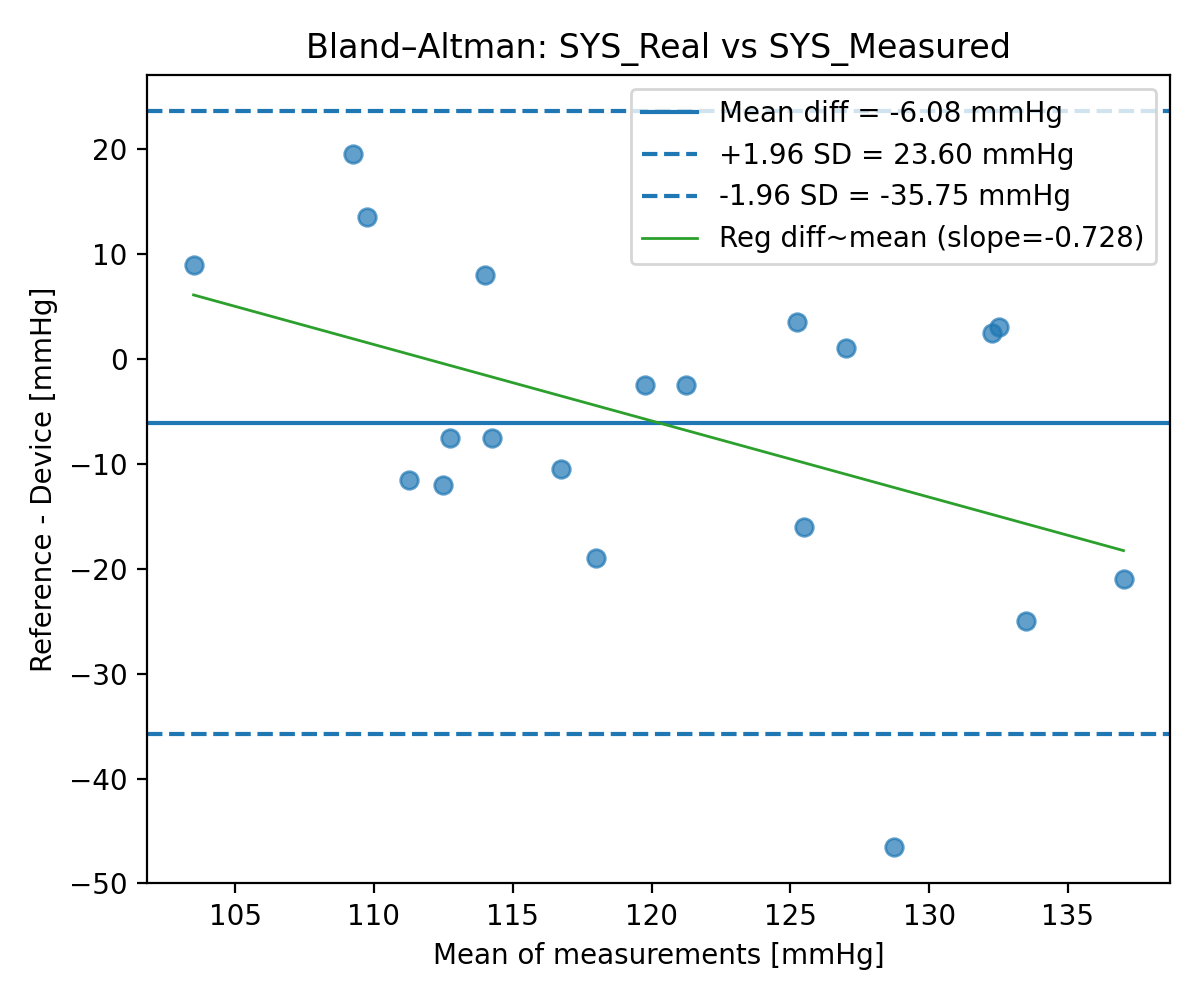}
         \caption{ }
         \label{fig:bland-altman-sys}
     \end{subfigure}
     \hfill
     \begin{subfigure}[b]{0.45\textwidth}
         \centering
         \includegraphics[width=\textwidth]{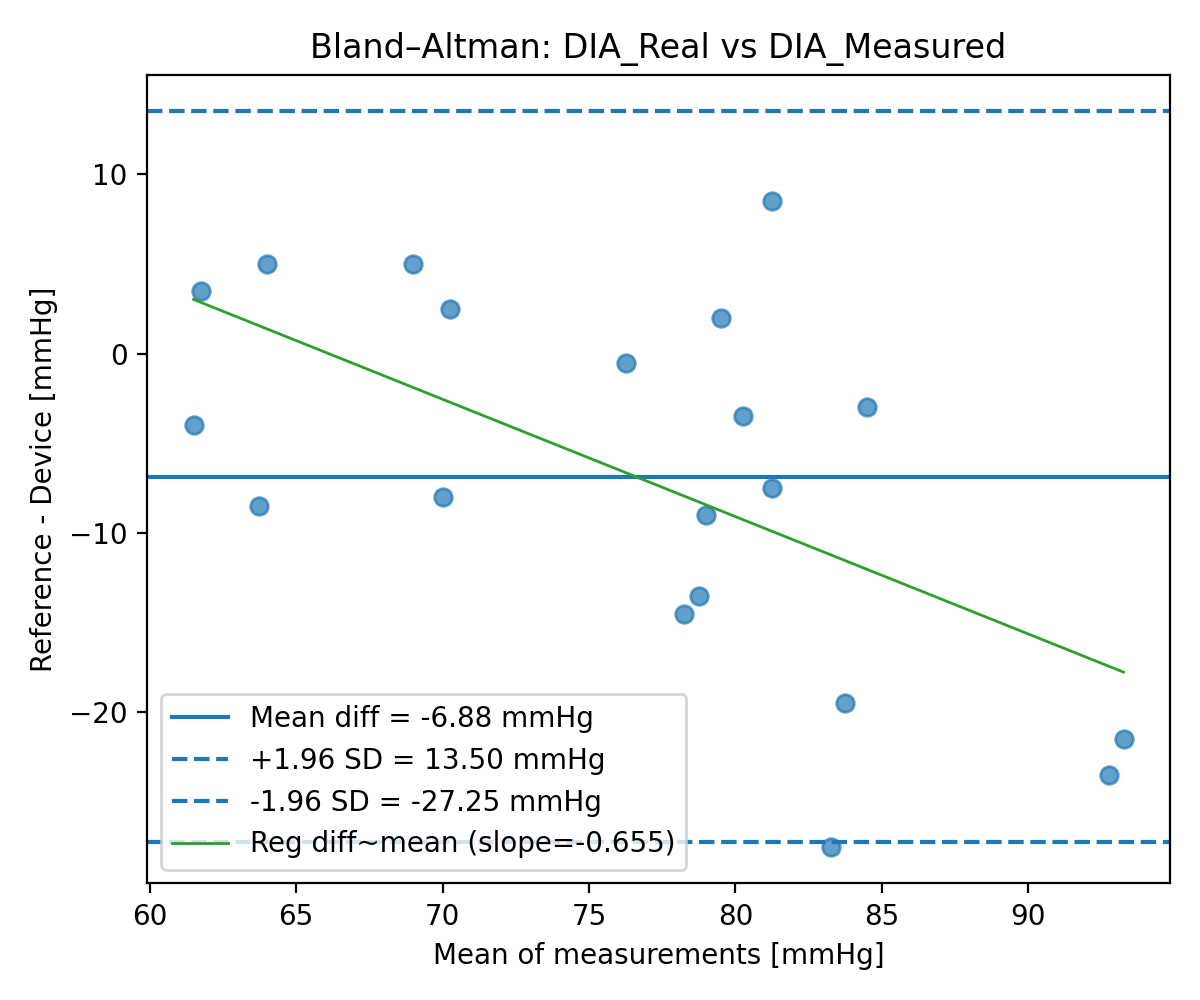}
         \caption{ }
         \label{fig:bland-altman-dia}
     \end{subfigure}
    \caption{Bland-Altman plots for the experiments of A) Pulse. B) Systolic pressure. C) Diastolic pressure.}
    \label{fig:bland-altman}
\end{figure}

The bias is defined as the discrepancy between the measurements obtained from the reference device and those derived by the system. Pulse rate is found to be -\qty{6.8}{\bpm} (or -\qty{4.21}{\bpm} in the absence of the outlier), while blood pressures are recorded at approximately -\qty{6}{\mmHg}. Consequently, the system has a propensity to underestimate the true measurement of pulse and pressure. This same trend is reflected in the Limits of Agreement (LoA) of the Bland-Altman plot, which predict the maximum expected error in 95\% of measurements, where the lower limits are slightly greater than the upper limits.

This discrepancy can be attributed primarily to the geometry and deformation capacity of the actuator. The softness and automatic grip of the actuator prevent it from surrounding the arms with the same firmness as a human would. This issue can be rectified, if deemed necessary, through the refinement of the MAP and systolic pressure detection algorithms outlined in Section~\ref{subsecc:signal-proc}.

A thorough examination of the magnitude of these values reveals a close resemblance to the specifications required for medical measuring devices. Specifically, the ISO 81060-2:2018 standard stipulates a bias of less than or equal to 5 units and an SD of less than or equal to 8 units. These margins are reached in pulse estimation, while pressure estimation remains marginally elevated. There are no direct specifications on LoA in this standard due to their relationship with bias and SD. A comparison of the results, in terms of LoA, with existing wearables\cites{Alfonso2022, loa-bpm} on the market reveals that those provided by our system are slightly higher than the others.

\subsubsection{Homoscedasticity Test}
Finally, homoscedasticity was studied to ascertain whether the system's errors are uniform across all pressure ranges. This phenomenon can be corroborated through the analysis of the Bland-Altman plots presented in Figure~\ref{fig:bland-altman}. These results indicate that the pulse measurement is clearly homoscedastic, although the green line exhibits a steep slope due to the influence of the outlier. Conversely, systolic pressure demonstrates a consistent variability in measurement outcomes across the entire spectrum of values. In contrast, diastolic pressure may exhibit a specific tendency.

To perform a more rigorous analysis, the Spearman's correlation coefficient ($\rho$) between the differences and the mean values has been calculated. Additionally, the p-value is obtained for this test. The detailed approach is delineated in Supplementary Note S5. The value of $\rho$ quantifies the correlation between the errors and the patient's actual pressure, with lower values being preferable. The p-value is a measure of the probability of obtaining this $\rho$ given the assumption that the data are homoscedastic. Consequently, high values are of interest.

The test showed no significant correlation for pulse rate without the outlier ($\rho$ = 0.09, p = 0.68), systolic pressure ($\rho$ = 0.08, p = 0.72), and diastolic pressure ($\rho$ = 0.11, p = 0.63), indicating homoscedasticity. A significant correlation was observed when including one outlier in the pulse data ($\rho$ = 0.44, p = 0.045), confirming that the outlier introduces proportional bias.

In summary, the findings of this section lend credence to the system's validity, as they demonstrate its precision and the consistent variability of errors across the entire measurement range for each variable. It is imperative to acknowledge that the objective of this study does not encompass the exact measurement of blood pressure and pulse, as if the system were intended for daily medical use. Conversely, it is the assessment of the patient's condition that facilitates the determination of the priority of the patient's removal by rescue teams.

\subsection{Validation in SAR Scenarios}

\subsubsection{Scenario Setup}

\begin{figure}[!ht]
    \centering
    \begin{subfigure}[b]{\textwidth}
        \centering
        \includegraphics[width=\linewidth]{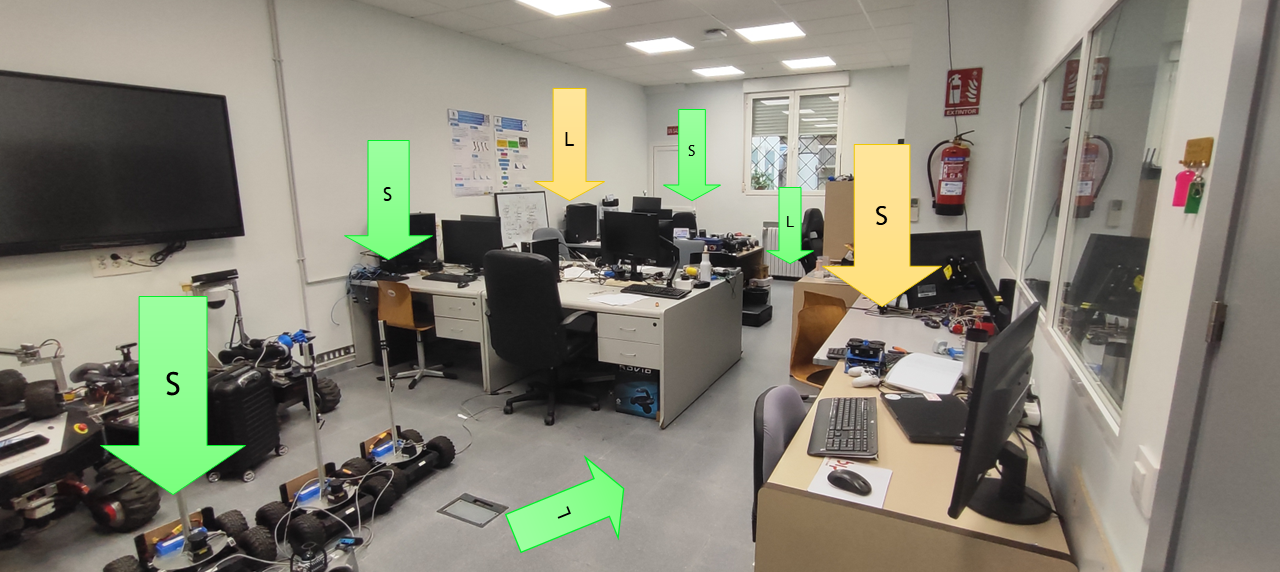}
        \caption{~}
        \label{fig:val-room}
    \end{subfigure}
    \hfill
    \begin{subfigure}[b]{0.45\textwidth}
        \centering
        \includegraphics[width=\linewidth]{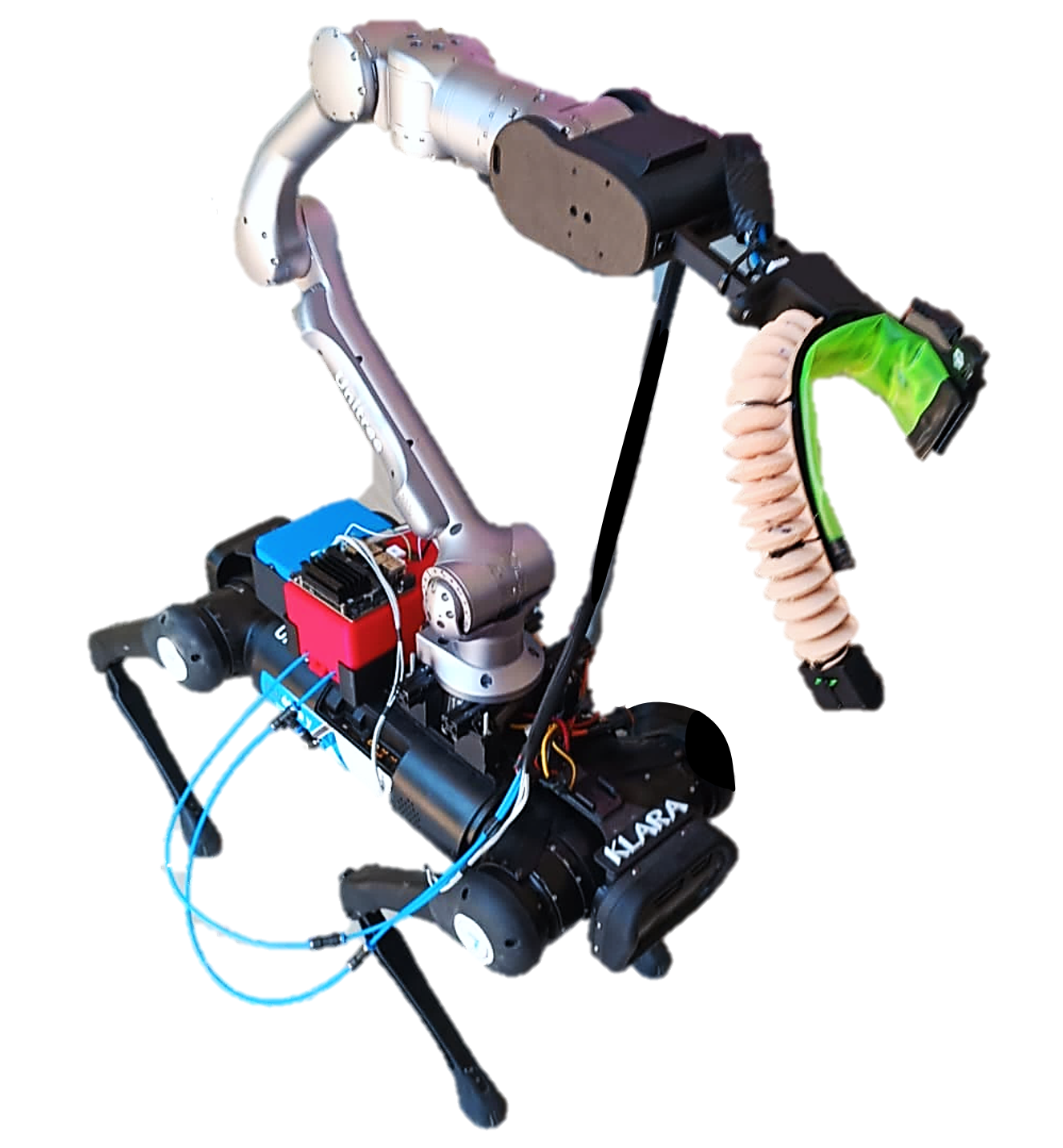}
        \caption{~}
        \label{fig:val-klara}
    \end{subfigure}
    \hfill
    \begin{subfigure}[b]{0.45\textwidth}
        \centering
        \includegraphics[width=\linewidth]{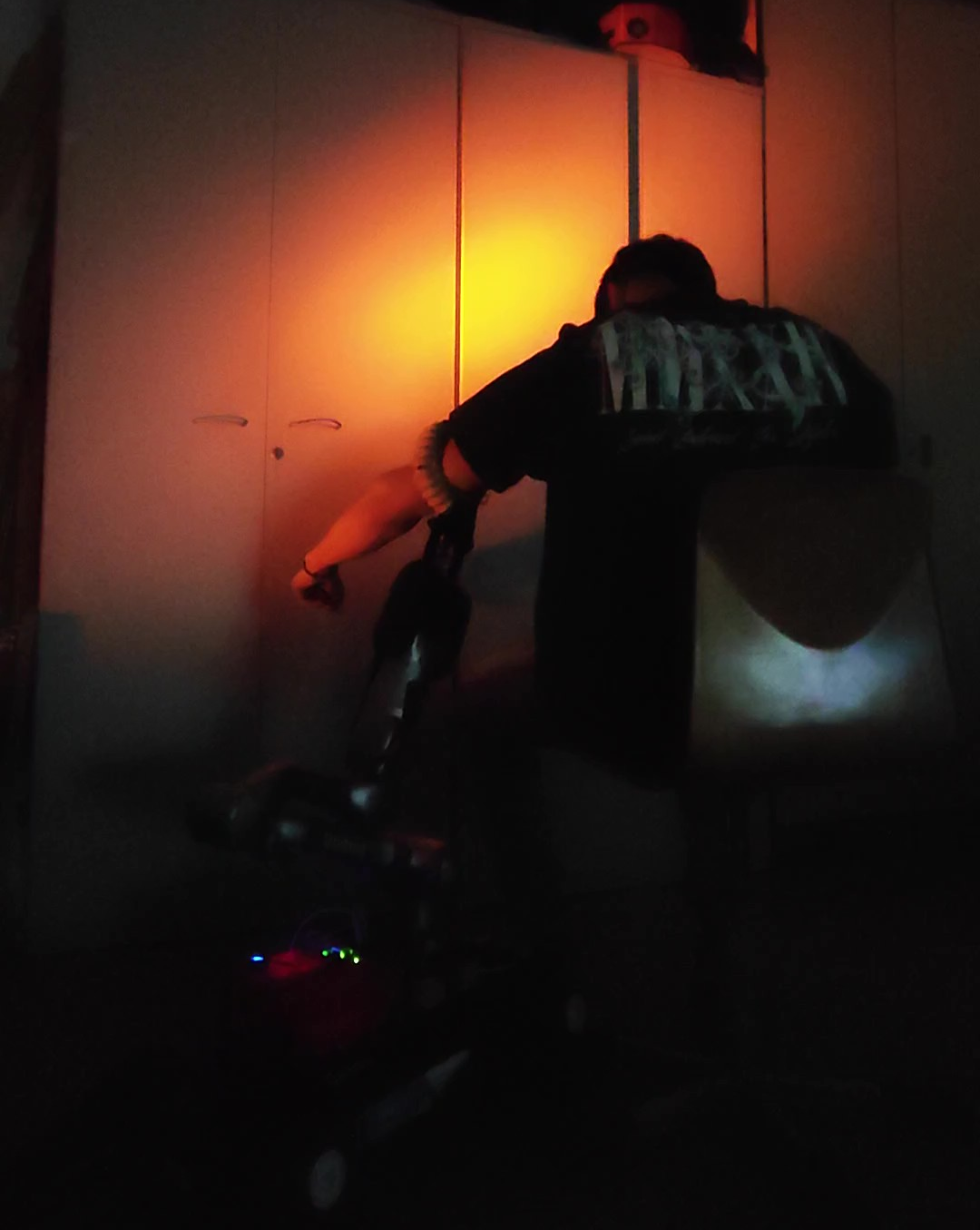}
        \caption{~}
        \label{fig:val-measuring}
    \end{subfigure}

    \caption{Validation scenario and tests: 
        a) Post-disaster scenario used for validation. Green arrows indicate the presence of a human victim, while yellow arrows indicate the presence of a mannequin. The position of the victim (seated or lying down) is specified by the letter inside the arrow.
        b) Complete system setup for validation. 
        c) System measuring vital signs to a victim.}
    \label{fig:val}
\end{figure}

After testing the gripper's ability to measure pulse and blood pressure with sufficient accuracy for rescue environments, we evaluated the entire system. For this evaluation, the gripper was attached to the robotic arm mounted on the quadruped described in Section~\ref{subsecc:robot}. A LIPO battery powered the arm, the Jetson Nano, and all of the system's electronics. The Aliengo has its own battery system, as previously presented. Figure~\ref{fig:val-klara} shows an image of the complete system.

For the design of the post-disaster validation scenario, NIST standards were followed\cites{Yanco2004}. This organization delineates three categories -yellow, orange, and red, from least to most difficult— based on the complexity of the scenario, thus facilitating a comparative analysis of results in robot research and competitions.  Given that the objective of the work does not center on mobility over very uneven terrain, it was decided to construct a yellow-type scenario. The experiment was conducted under low-light conditions, simulating a fire, to prove the strength of the system over pulse measurement methods based on visual PPG, which present limitations in such conditions.

The testing scenario comprised a chamber measuring 9 by 4 square meters, as depicted in Figure~\ref{fig:val-room}. The simulation of fire conditions was conducted by employing a smoke machine, with the intensity of light measured at \qty{50}{\lux} in all zones situated more than one meter from the machine. A total of five subjects and two mannequins were situated within the designated chamber. Two of the subjects and one of the mannequins were lying down, while the rest were seated.

The system was teleoperated by a human operator who had experience in this area. Its main objective was to measure the vital signs of all victims without causing them discomfort. For each subject, the system maneuvered the quadruped until it was in proximity, at which point it operated the arm until the gripper made contact with the human arm. Subsequently, a start command was transmitted to the Jetson Nano, thereby initiating the autonomous measurement process. This entailed the closure of the clamp, the application of pressure to the cuff, and the estimation of heart rate and pressure. Subsequent to the conclusion of the process, the pulse and pressure of the human subjects were measured using the aforementioned commercial sphygmomanometer to assess their actual conditions. A picture of a measurement is depicted in Figure~\ref{fig:val-measuring}.

\subsubsection{Validation Results}

The duration of the complete test was 18 minutes and 12 seconds, which equates to an average of 2 minutes and 36 seconds per subject, whether actual or a mannequin. This duration is deemed reasonable, considering the necessity of measuring the pulse, which requires almost one minute, and the requirement for the operation of the arm to be executed slowly to ensure precision and prevent injury to the subject.

For both mannequins, the system returns an error message to the operator, indicating that no pulse or pressure was detected, as would be expected. For the human subjects, the MAE for pressure was \qty{9.73}{\bpm} for the pulse, \qty{10.13}{\mmHg} for the systolic pressure, and \qty{11.02}{\mmHg} for the diastolic pressure. The errors are similar to those previously obtained in the Performance Analysis. This suggests that, with proper system operation, access to the victim's arm by the system is feasible, enabling the measurement of pulse and pressure without encountering any significant impediments.  Therefore, it can be posited that the system's performance remains uncompromised in a post-disaster scenario.

In addition, an analysis of the safety perceptions of the victims themselves as passive users of the system has been conducted. Although it is possible that actual users in a post-disaster scenario may be unaware, the objective was not to obtain feedback from assisted individuals but rather to investigate whether soft robotics can leverage their compliance to enhance human safety.

Subsequently, all participants were requested to evaluate the extent of their fear, using a 5-point scale, with 1 indicating no fear and 5 indicating high levels of fear. Two reported a fear of 2, others of 3, and the remaining one of 4. In consequence, the mean level of fear experienced by victims was 2.8, which is slightly lower than the midpoint on the scale. It is believed that the utilisation of a TPU with a low hardness value serves to alleviate the adverse sensations that may be experienced when in contact with an iron manipulator.

It is evident that none of the participants rated their experience as maximally fearful, with only one individual selecting level 4. Informal discussions with those who selected ratings 3 and 4 indicated that these were attributable to prior unfavourable experiences with robots, a perceived manipulation speed that remained excessive, and a deficiency in confidence in the safety systems in the event of operator performance failure. This suggests that the system is not only safe for victims but they also perceive it as such. Moreover, these findings do not preclude the potential application of these results in the development of future research line. 

\section{Conclusion}

Despite the development of numerous robotic systems designed to evaluate and even perform triage of victims in post-disaster scenarios, none of these systems have been capable of measuring pressure. Nonetheless, this variable can facilitate a more profound comprehension of the victim's condition, forecast issues that have not been discerned through the utilization of alternative variables, and, when monitored over a certain period of time, permit a more meticulous examination of the injured individual's progression over time. To solve this, in this work, we presented a robotic system capable of measuring heart pulse and arterial pressure.

The system is composed of a teleoperated quadruped robot with a six-degree-of-freedom arm mounted over it, a soft gripper that enters contact with the victim's arms, a pneumatic and portable system, and a central computer that enables the system to autonomously obtain vital constants. Upon contact between the gripper and the victim, the clamp is activated, enveloping the arm. This action initiates a process analogous to that of sphygmomanometers, leading to the inflation of a cuff within the gripper. This inflation results in the estimation of vital signs, including heart rate and blood pressure. A set of algorithms has been developed to accurately identify all the peaks in the pulse signal. Portability, still an ongoing challenge in soft robotics, has been addressed with success. Since the arm of the robot has reduced its weight-carrying capabilities, a system based on DC motors and a pump has been developed to mount a complete pneumatic system over a robot. 

A first validation has been conducted that exclusively addressed the capabilities of the gipper measurement. A substantial number of experiments have been executed and a range of statistical metrics have been obtained. The median errors remained below \qty{10}{\mmHg}, indicating that large outliers were rare and that the sensing principle provides a reliable baseline estimation. Although the MAE values are marginally elevated relative to the expected values for medical devices, they remain within acceptable limits and are sufficient to ascertain the condition of the victim in post-disaster scenarios where precise results are not imperative.

A thorough examination of the bias and limits of agreement reveals that the system exhibits a tendency to underestimate the true measurements of pulse and pressure. This phenomenon is likely attributable to soft material deformation dynamics or calibration misadjustments. Signal processing improvements have been demonstrated to be a viable solution to this problem. Furthermore, the homoscedasticity test demonstrates that system errors are consistent across the range of pulse and pressure values.

Finally, the complete system has been tested on a SAR scenario with poor luminosity conditions. In this phase, the efficacy of the system was demonstrated not only by its precision but also by its capacity to remotely manipulate the robot's arm and the arm's proximity to the victims. The operation took a reasonable amount of time. As humans and mannequins were combined in this scenario, it was observed that the system did not detect a pulse in mannequins that did not have a pulse, and that people did not experience high levels of fear when faced with the measure. Future research will entail the mitigation of bias, the modification of the arm design to enhance safety, and the incorporation of prior developments related to teleoperation.

\clearpage
\section{Experimental Section}

\threesubsection{Notes on the Performance Analysis Experiments}\\

To analyze the device's ability to measure pulse and blood pressure, 20 patients between the ages of 21 and 58 were evaluated. Supplementary Figures S5-S8 show, respectively, the distribution of the sample by gender, age, diastolic pressure, and systolic pressure. The sample was taken as uniformly as possible. The participants in the study asserted that they had no prior medical records indicating the presence of heart disease or related conditions.

The sampling procedure was standardized in order to minimize variability associated with the environment or patient behavior. Each volunteer remained seated, with their left arm extended and uncovered, resting on a table. The patient was asked to remain as still as possible, avoiding both conversation and any unnecessary movement during the measurement. The standardized posture is depicted in Supplementary Figure S9.

The measurement was taken using only the gripper, not using the complete robotic system. When the patient was in the correct position, an operator brought the gripper closer, leaving it open. At that moment, a second operator started the system. The actuator was then automatically closed and then the pulse and blood pressure were measured. This process took less than 1 minute. After a 3-minute wait, heart rate and blood pressure were measured using a commercial Omron M2+ sphygmomanometer, as there were no operators specialized in the use of acoustic sphygmomanometers.

This process was performed twice per person, with a 5-minute interval between each measurement, allowing the arm to recover its vascular tone and ensuring that subsequent readings were not affected. 

Data from the experiments are reported in Supplementary Table S2.

\medskip
\textbf{Supporting Information} \par 
Supporting Information is available from the Wiley Online Library or from the author.

\medskip
\textbf{Acknowledgments} \par 
This work is the result of research activities carried out at the Centre for Automation and Robotics, CAR (UPM-CSIC), in the facilities of the Escuela Técnica Superior de Ingenieros Industriales, within~the Robotics and Cybernetics research group (RobCib). This work is supported by “Ayudas para contratos predoctorales para la realización del doctorado con mención internacional en sus escuelas, facultad, centros e institutos de I+D+i”, which is funded by “Programa Propio I+D+i 2022 from Universidad Politécnica de Madrid” and~by the “Proyecto CollaborativE Search And Rescue robots (CESAR)” (PID2022-142129OB-I00) funded by MCIN/AEI/10.13039/501100011033 and “ERDF/EU”

\medskip
\textbf{Conflict of Interest} \par
The authors declare no conﬂict of interest.

\medskip
\textbf{Author Contributions} \par
J.F.G.: Conceptualization, Methodology, Formal Analysis, Investigation, Data Curation, Writing - Original Draft, Writing - Review \& Editing, Visualization. C.C.U.: Conceptualization, Formal Analysis, Investigation, Resources, Writing - Original Draft, Writing - Review \& Editing, Supervision. . A.S.: Methodology, Software, Validation, Formal analysis, Investigation, Data Curation. J.C.: Conceptualization, Writing - Review \& Editing, Supervision, Project administration, Funding acquisition. A.B.: Conceptualization, Writing - Review \& Editing, Supervision, Project administration, Funding acquisition.

\medskip
\textbf{Data Availability Statement} \par
The data that support the ﬁndings of this study are available in the supplementary material of this article.

The following repository contain all the code for the teleoperation of the gripper, the signal processing algorithms and the data and results of the Performance Analysis:  \\ \url{https://github.com/Robcib-GIT/Soft-Actuator-Measurement/}

\medskip

%
\bibliographystyle{MSP}
\bibliography{bibliography/references, bibliography/SoftRobotics}







\end{document}